\documentclass{article}

\usepackage{arxiv}

\usepackage[utf8]{inputenc} 
\usepackage[T1]{fontenc}    
\usepackage{hyperref}       
\usepackage{url}            
\usepackage{booktabs}       
\usepackage{amsfonts}       
\usepackage{nicefrac}       
\usepackage{microtype}      
\usepackage{lipsum}
\usepackage{graphicx}
\usepackage{amsmath}
\usepackage{adjustbox}
\usepackage{multirow}
\usepackage{tabularx}
\usepackage{indentfirst}

\graphicspath{ {./images/} }

\title{Advancing Facial Stylization through Semantic Preservation Constraint and Pseudo-Paired Supervision}

\author{
 Zhanyi Lu \\
  School of Electrical and Electronic Engineering\\
  University of Shanghai Jiao Tong University\\
  \texttt{luzhanyi@sjtu.edu.cn} 
 \And 
 Yue Zhou \\
  School of Electrical and Electronic Engineering\\
  University of Shanghai Jiao Tong University\\
  \texttt{zhouyue@sjtu.edu.cn} 
}

\begin{document}

\maketitle
\begin{figure}[ht]
\includegraphics[width=1.0\textwidth]{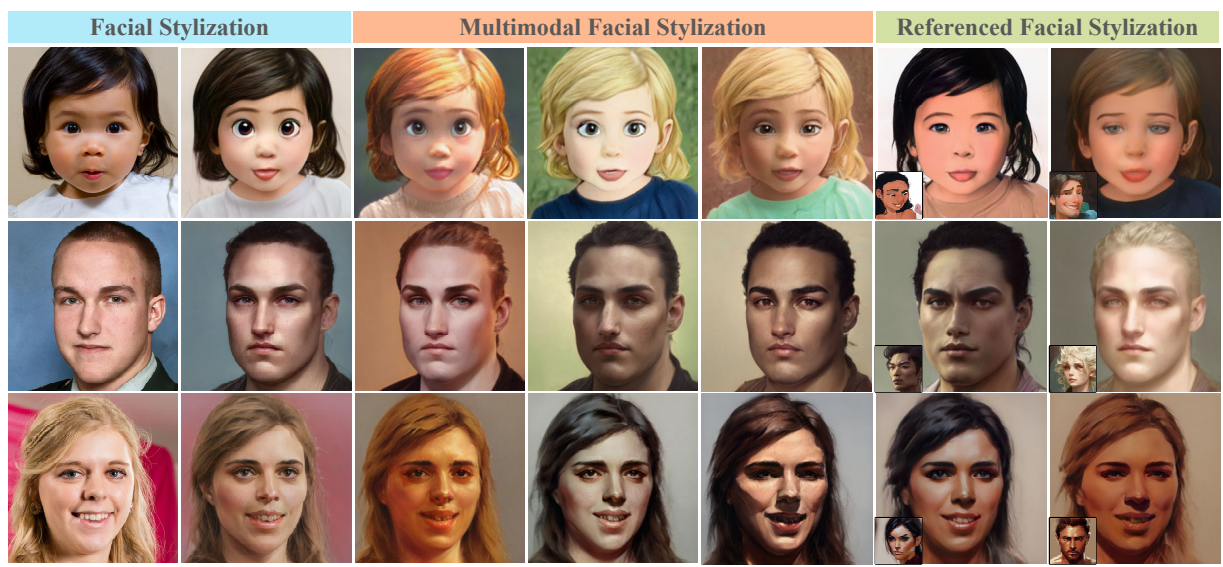}
\caption{We propose a facial stylization approach supporting general, multimodal, and reference-guided stylization. The styles shown are \textit{cartoon}, \textit{fantasy}, and \textit{impasto}, respectively.}
\label{fig}
\end{figure}

\begin{abstract}
Facial stylization aims to transform facial images into appealing, high-quality stylized portraits, with the critical challenge of accurately learning the target style while maintaining content consistency with the original image. Although previous StyleGAN-based methods have made significant advancements, the generated results still suffer from artifacts or insufficient fidelity to the source image. We argue that these issues stem from neglecting semantic shift of the generator during stylization. Therefore, we propose a facial stylization method that integrates semantic preservation constraint and pseudo-paired supervision to enhance the content correspondence and improve the stylization effect. Additionally, we develop a methodology for creating multi-level pseudo-paired datasets to implement supervisory constraint. Furthermore, building upon our facial stylization framework, we achieve more flexible multimodal and reference-guided stylization without complex network architecture designs or additional training. Experimental results demonstrate that our approach produces high-fidelity, aesthetically pleasing facial style transfer that surpasses previous methods.
\end{abstract}


\section{Introduction}
Facial stylization has become a bridge between real and virtual worlds. It automatically renders real facial images into artistic styles, such as cartoons or anime, providing users with new forms of self-expression and enhancing interactive experiences on digital platforms \cite{cartoon_survey}. Its main challenges are achieving high visual quality, good aesthetics, and preserving the original identity.

While image-to-image (I2I) translation methods \cite{cyclegan, ugatit, GNR} have advanced facial stylization, they often demand substantial training resources and struggle to produce high-quality outputs. In contrast, StyleGAN \cite{stylegan, stylegan2} excels at generating high-resolution facial images and can adapt to new styles with limited data \cite{stylealign}. StyleGAN-based facial stylization methods \cite{toonify, UI2I-style, agliegan, dualstylegan} involve mapping a real image into the latent space of a pretrained StyleGAN model and then decoding it with a finetuned style-specific model. Although these methods yield high-quality visuals, they occasionally introduce artifacts and lack fidelity in preserving the original content.

Diffusion Models (DMs) have made substantial advancements in text-to-image generation \cite{glide,ldm} and have been applied to various image-to-image translation tasks such as style transfer \cite{diffusioninstyle, stylediffusion, inst, styleid}. However, for portrait stylization, we opt not to use diffusion-based models due to the following concerns: first, although pre-trained DMs perform admirably in tasks that bridge textual and visual domains, when handling purely visual tasks like portrait stylization, current methods struggle to generate the necessary geometric deformations or texture simplifications in portraits; second, compared to GAN-based methods, pre-trained DMs possess more complex structures with a larger number of parameters, leading to slower stylization process; last, we believe that there are still effective improvements in StyleGAN-based facial stylization. These perspectives have been validated through experiments conducted in this paper.

We argue that previous StyleGAN-based methods have overlooked the semantic alterations in StyleGAN's latent space caused by changes in latent distribution during finetuning, which reduces output quality and fidelity. To address this, we propose two key enhancements: first, a semantic preservation constraint to maintain essential semantics during finetuning; second, the use of pseudo-paired supervision, involving the creation of a multi-level pseudo-paired dataset and paired supervision to mitigate data distribution shifts, thus preserving content correspondence between real and portrait domains. These enhancements result in higher quality and more faithful stylization.

Additionally, users may desire stylized portraits that match a specific reference image or require diverse outputs with varying degrees of stylization. Leveraging StyleGAN's latent space and inversion methods, our method supports not just general stylization, but also controllable multimodal and reference-guided stylization, offering enhanced flexibility without requiring specialized network design or additional training. Our results are illustrated in Figure \ref{fig}. More results are available in the supplementary material.

In summary, this paper proposes a facial stylization method with the following contributions:

First, we introduce a StyleGAN-based facial stylization approach augmented with semantic preservation constraint and pseudo-paired supervision, which generates high-quality and faithful stylized portraits.

Second, we present a method for creating multi-level pseudo-paired data from stylized portraits, resulting in pseudo-realistic face images with varying degrees of authenticity.

Lastly, our approach enables general, flexible multimodal and reference-guided stylization without additional network design or training, enhancing user experience.

\section{Related Works}
\label{sec:related works}

\subsection{Facial Stylization with GANs}
Facial stylization is an application of image-to-image translation where GAN-based methods \cite{GANs,dcgan} have made significant strides. Pix2Pix \cite{pix2pix} effectively translates images using conditional GANs \cite{cgan} but relies on paired data. CycleGAN and similar approaches \cite{cyclegan, dualgan, DiscoGAN} introduced unsupervised learning to remove the paired data requirement. Subsequent advancements \cite{ugatit, GNR} have improved detail handling and conversion fidelity. However, these models often struggle with learning complex bidirectional mappings from scratch, leading to suboptimal visual quality.

StyleGAN \cite{stylegan, stylegan2} is known for high-quality realistic faces and effective fine-tuning capabilities \cite{stylealign}. StyleGAN-based methods \cite{toonify,agliegan,UI2I-style,dualstylegan} encode images into latent space and use a finetuned StyleGAN as a decoder, enhancing image quality without the need for complex mapping networks. Toonify \cite{toonify} combines high-resolution layers from the fine-tuned model with low-resolution layers from the pre-trained model to achieve effective stylization. UI2I-Style \cite{UI2I-style} introduces noise or encodings into generator through layer swapping to support multimodal and reference-guided stylization. DualStyleGAN \cite{dualstylegan} adds an external path to module reference-guided translation. However, issues such as artifacts and poor fidelity still exist. 

We attribute these issues to the underestimation of semantic changes during StyleGAN finetuning. To address this, we introduce a semantic preservation constraint and pseudo-paired supervision to enhance model correlation and improve image quality. Additionally, our method achieves controllable multimodal and reference-based facial stylization without requiring additional network design or training.

\subsection{Style Transfer with Diffusion Models}
Style transfer aims to render content images into specific artistic styles. Gatys et al. \cite{nst} pioneered neural style transfer using pre-trained CNNs, and subsequent studies have enhanced real-time performance\cite{texture, perceptual} and transferring with arbitrary reference images \cite{adain,adaain,dynamic}. To improve stylization effects, models have evolved from CNNs to flow-based models \cite{artflow} and transformers \cite{styleformer,stytr2}.

Recently, diffusion models have achieved significant breakthroughs in text-to-image generation due to their powerful feature representation capabilities \cite{ldm}. These models have also been applied to I2I tasks such as style transfer. VCT \cite{vct} extracts embeddings from the source and reference images using a content-concept inversion process and integrates them with a content-concept fusion process. InST \cite{inst} introduces style encodings from reference images through text inverse transformation; StyleID \cite{styleid} replaces content representations with style information within the attention layer; NTC \cite{ntc} employs diffusion models for cartoon rendering using image and rollback disturbance.

Despite their advancements, diffusion-based stylization still faces limitations in generating complex or abstract styles or preserving content due to the loss of direct utilization of textual prompts. Also, they are primarily suited for reference-guided tasks. For pure visual tasks like portrait style transfer, we utilize pre-trained StyleGAN to better capture facial features, our method also allows general and multi-modal stylization besides referenced stylization.

\subsection{GAN Inversion}
GAN inversion refers to embedding a given image into the latent space of a pretrained GAN to obtain an encoding that accurately reconstructs the image \cite{gansurvey}. StyleGAN's latent space is rich in semantic information, enabling image editing via latent manipulations \cite{interpreting, ganspace, stylespace, styleclip}.

StyleGAN possesses a \(Z\) latent space with simple distribution, which is transformed into the semantically informative $W$ space by its mapping network. The \(Z^+\) space \cite{agliegan} extends the \(Z\) space with finer details, while \cite{UI2I-style} enhances \(W\) space reconstruction by learning an indirect \(V\) space \cite{sefa}. The \(W^+\) space \cite{image2stylegan, image2stylegan++} extends \(W\) with greater expressiveness but reduced editability. GAN inversion techniques include optimization-based methods \cite{image2stylegan, image2stylegan++, stylegan2}, which are computationally expensive but precise, and encoder-based methods \cite{psp, e4e, restyle}, which are faster but less precise.

In this paper, we use a modified pSp encoder \cite{psp} to map facial images into StyleGAN's $W$ space to ensure real-time performance. For reference-guided facial stylization, an optimization-based method \cite{UI2I-style} is used to obtain $W$ space encodings, allowing storage and reuse, thereby minimizing redundant computation.

\section{Method}
\label{sec:method}
\begin{figure*}[t]
\centering
\includegraphics[width=.9\textwidth]{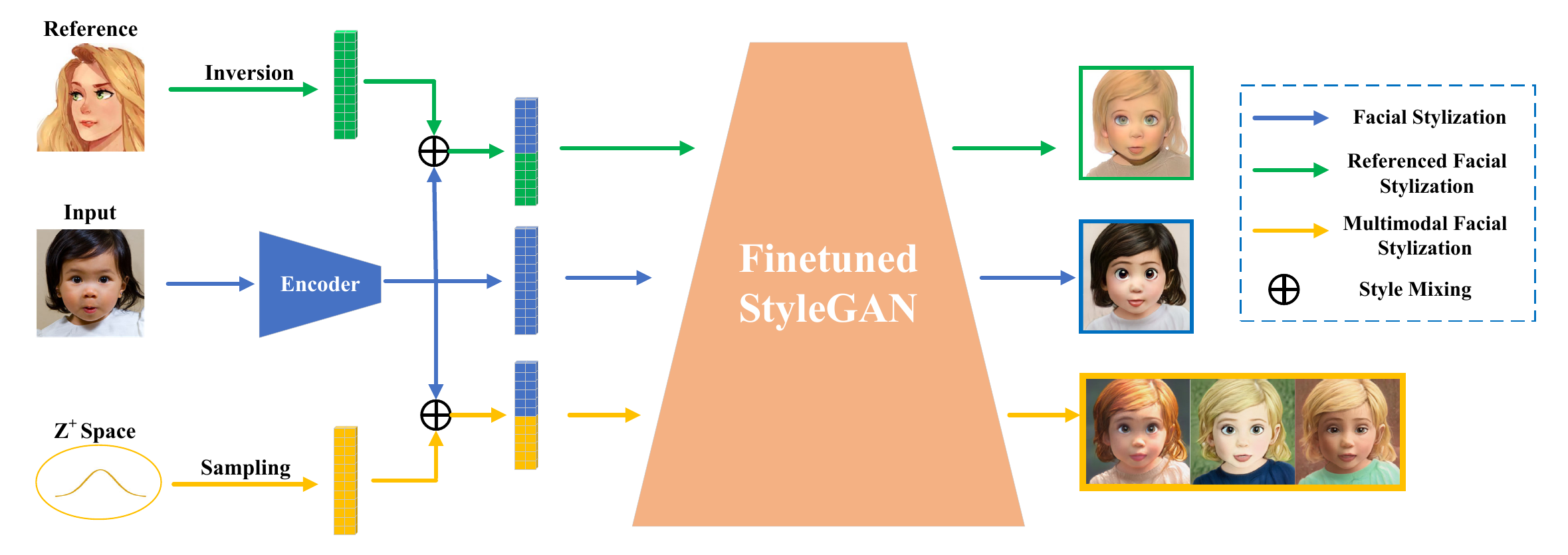} 
\caption{Facial stylization framework: supporting reference-guided and multimodal facial stylization} 
\label{framework}
\end{figure*}

Figure \ref{framework} illustrates the proposed facial stylization framework. We use an adjusted pSp encoder \cite{psp} to embed input images into the $W$ space of a StyleGAN pretrained on the FFHQ dataset \cite{stylegan}, and employ a finetuned StyleGAN with our semantic and pseudo-paired constraints as the decoder. Beyond conventional facial stylization, we achieve controllable multimodal and reference-guided facial stylization by mixing input encodings with random noise or reference image embeddings. Notably, our method requires minimal datasets and computational resources for finetuning StyleGAN, avoiding complex network designs and lengthy training processes.

In this section, we first explain the motivation and implementation of semantic and pseudo-paired constraints. Then, we describe the construction of multi-level paired data for pseudo-supervision training. Finally, we introduce the approach for achieving multimodal and reference-guided portrait stylization.

\subsection{Semantic and Pseudo-Paired Constraints}

\begin{figure}[!h]
\centering
\includegraphics[width=0.6\textwidth]{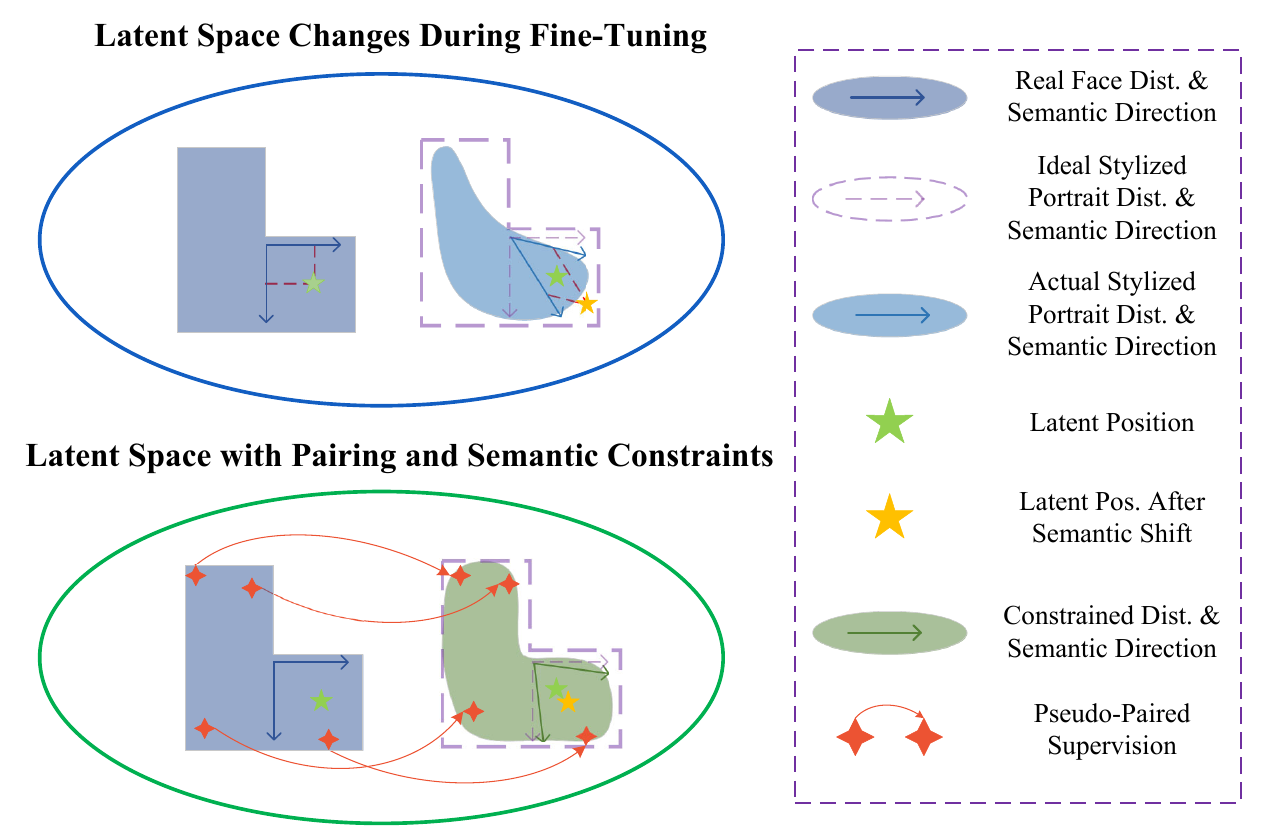} 
\caption{Changes in the latent space data distribution and semantics during finetuning. Top: Unconstrained; Bottom: Semantic and paired supervision constraints applied. } 
\label{changes}
\end{figure}

Figure \ref{changes} illustrates the changes in the distribution of the $W$ latent space during finetuning. Ideally, the latent and necessary semantic directions for stylized portraits should align with those of real faces to ensure content correspondence. During finetuning, the stylized portrait dataset provides specific style representations, while the pretrained model maintains content representations. However, style datasets used in finetuning are often much smaller than the FFHQ dataset. As training progresses, the model tends to overfit the style dataset, causing a shift in the learned latent distribution. This shift leads to two negative impacts: first, it increases the risk of mode collapse, degrading latent space interpolation and lowering image quality; second, it alters the semantic direction within the $W$ space, leading to inconsistencies in content expression between the original image and its stylized output from the same latent.

To address the aforementioned issues, we propose two key constraints during the fine-tuning process: semantic preservation constraint and pseudo-paired supervision. The semantic preservation constraint aims to maintain the essential semantics of the original domain when the generator incorporates a new style. This helps mitigate shifts in necessary semantics by utilizing reliable evaluation models \cite{perceptual-loss,arcface} to capture the semantic information in portraits. Additionally, if paired portrait data and their corresponding encodings can be obtained, overfitting can be reduced by aligning latent representations across domains, thereby creating a form of pseudo-paired supervision. As shown in Figure \ref{changes}, this constraint guides the reduction of distribution collapse with explicit paired data, improving image quality and fidelity.

\begin{figure}[h]
\centering
\includegraphics[width=0.65\textwidth]{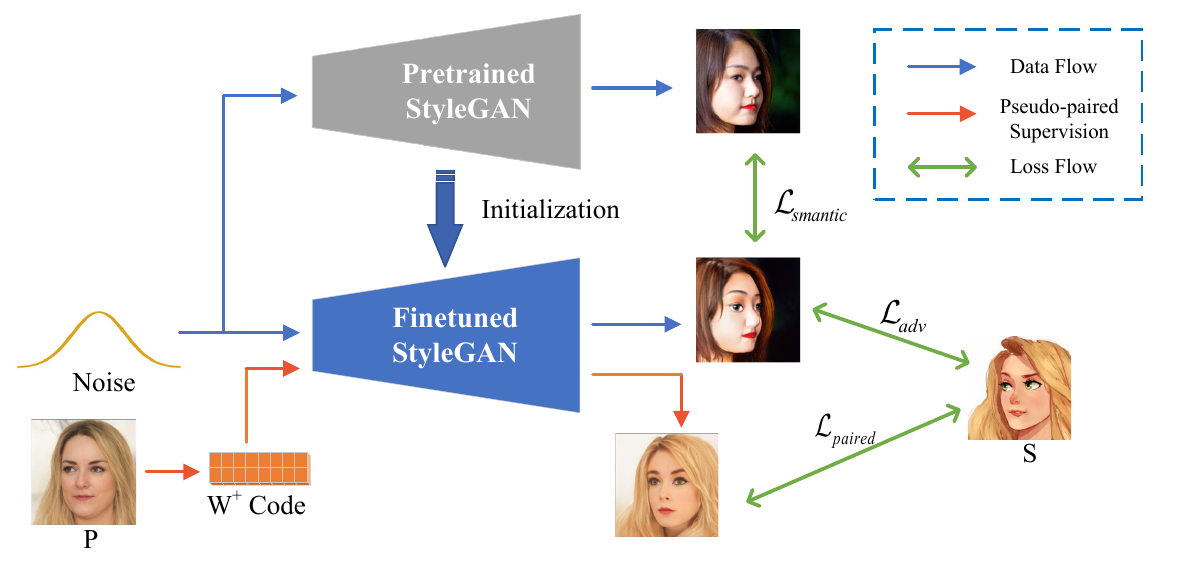} 
\caption{Model finetuning with semantic preservation constraint and pseudo-paired supervision.} 
\label{finetune}
\end{figure}

Figure \ref{finetune} shows finetuning process, we initialize with pretrained weights and compute adversarial loss between generated and style images to learn the target style. Simultaneously, the semantic preservation constraint is calculated by comparing the outputs of the pretrained and finetuned models using the same noise input. Pseudo-paired supervision is derived from the encodings of pseudo-paired data and their corresponding style images.

\noindent \textbf{Semantic Preservation Constraint}
We implement the semantic preservation constraint by calculating LPIPS\cite{lpips} and identity loss\cite{arcface} between the generated and source images to preserve necessary semantics. Let $G$ represent the pretrained model, $G^{\prime}$ the finetuned model and $z$ the random sampled noise, the constraint is defined as:
\begin{equation}
\begin{aligned}
\mathcal{L}_\mathrm{semantic}(G(z),G^{\prime}(z)) = \mathcal{L}_\mathrm{LPIPS}\left(G(z) - G^{\prime}(z)\right) \\
+ \lambda_\mathrm{ID}\mathcal{L}_\mathrm{ID}\left(G(z) - G^{\prime}(z)\right).
\end{aligned}
\end{equation}

\noindent \textbf{Pseudo-Paired Supervision}
Pseudo-paired supervision is applied using pseudo-paired data $(P, S)$ and their corresponding encodings $w^+$ (see Section 3.2). This constraint introduces supervised signals during finetuning to preserve the latent distribution, ensuring diversity and consistency in semantics between source and generated images.
\begin{equation}
\mathcal{L}_{paired}=\mathcal{L}_{\mathrm{LPIPS}}\left(G’(w^+)-S\right).
\end{equation}

Combining the adversarial loss during finetuning, the total loss is denoted as:
\begin{equation}
\mathcal{L}_{total}=\mathcal{L}_{adv}+\lambda_{sematic}\mathcal{L}_{sematic}+\lambda_{paired}\mathcal{L}_{paired}.
\end{equation}

\begin{figure}[htbp]
\centering
\includegraphics[width=0.5\textwidth]{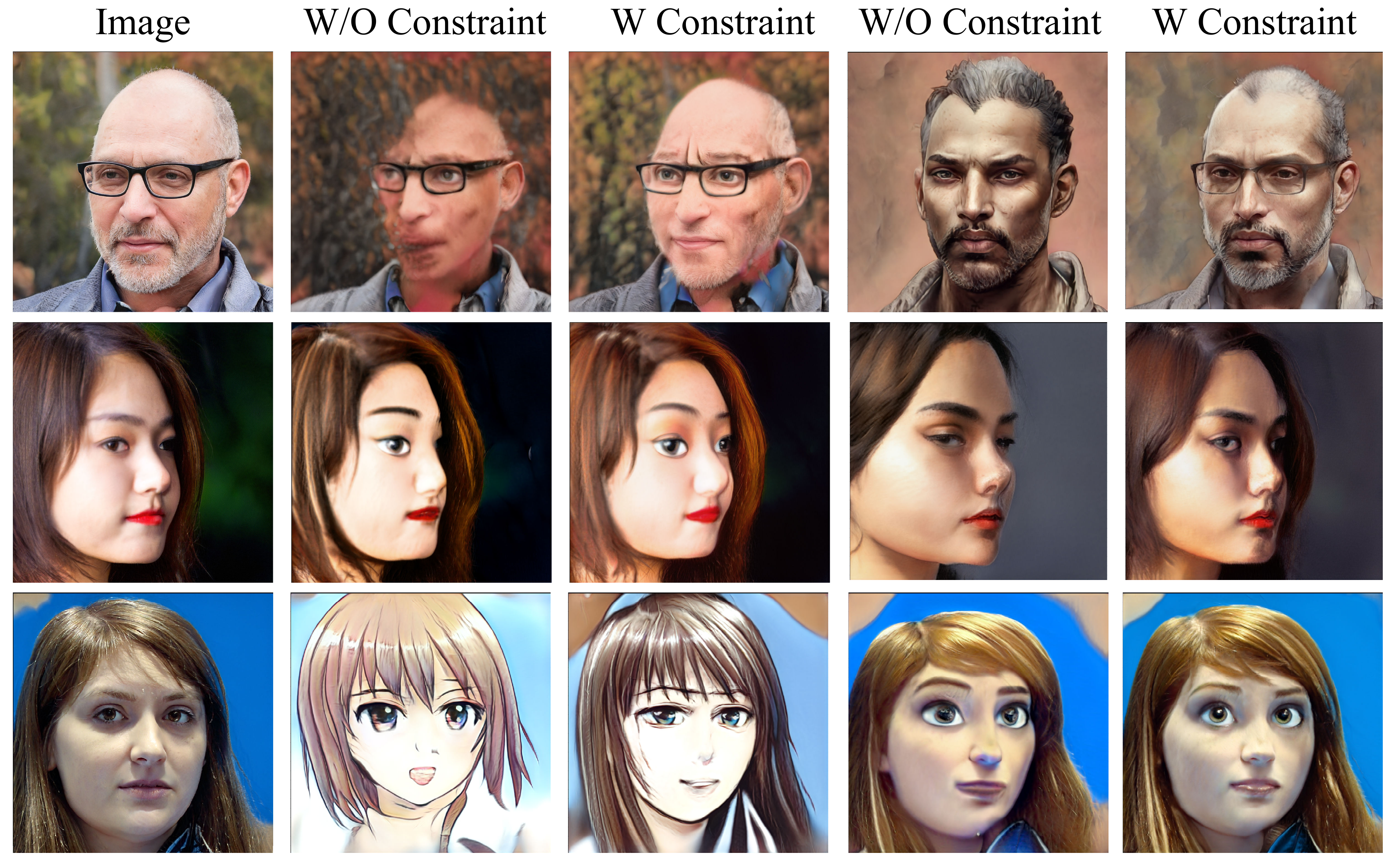} 
\caption{Results of random sampling images during finetuning with and without proposed constraints.} 
\label{compare_ft}
\end{figure}
Figure \ref{compare_ft} compares the results of randomly generated images with and without the proposed constraints under the same number of training iterations, showing that using the proposed constraints can better learn the target style while preserving the content features of the source image (such as facial structure, orientation, glasses and identity). This indicates that our method can alleviate mode collapse and maintain the semantic characteristics of the original domain.
\subsection{Multi-Level Pseudo-Paired Data Generation}


Supervised constraint requires paired data. By leveraging the properties of StyleGAN's latent space, we can generate multi-level pseudo-paired data. Assuming no semantic shift occurs, the semantic distributions of real and stylized portraits should be consistent: given a semantic code $ w^+ $, both the real portrait $P$ and the stylized portrait $S$ generated from $ w^+ $ should maintain content correspondence. Thus, an ideal content encoder trained on real portraits could map a stylized portrait $S$ back to semantic code $ w^+ $, allowing for the generation of a pseudo-real portrait $P$ by encoding $S$ and decoding it with generator $G$.

\begin{figure}[ht]
    \centering
    \includegraphics[width=0.65\textwidth]{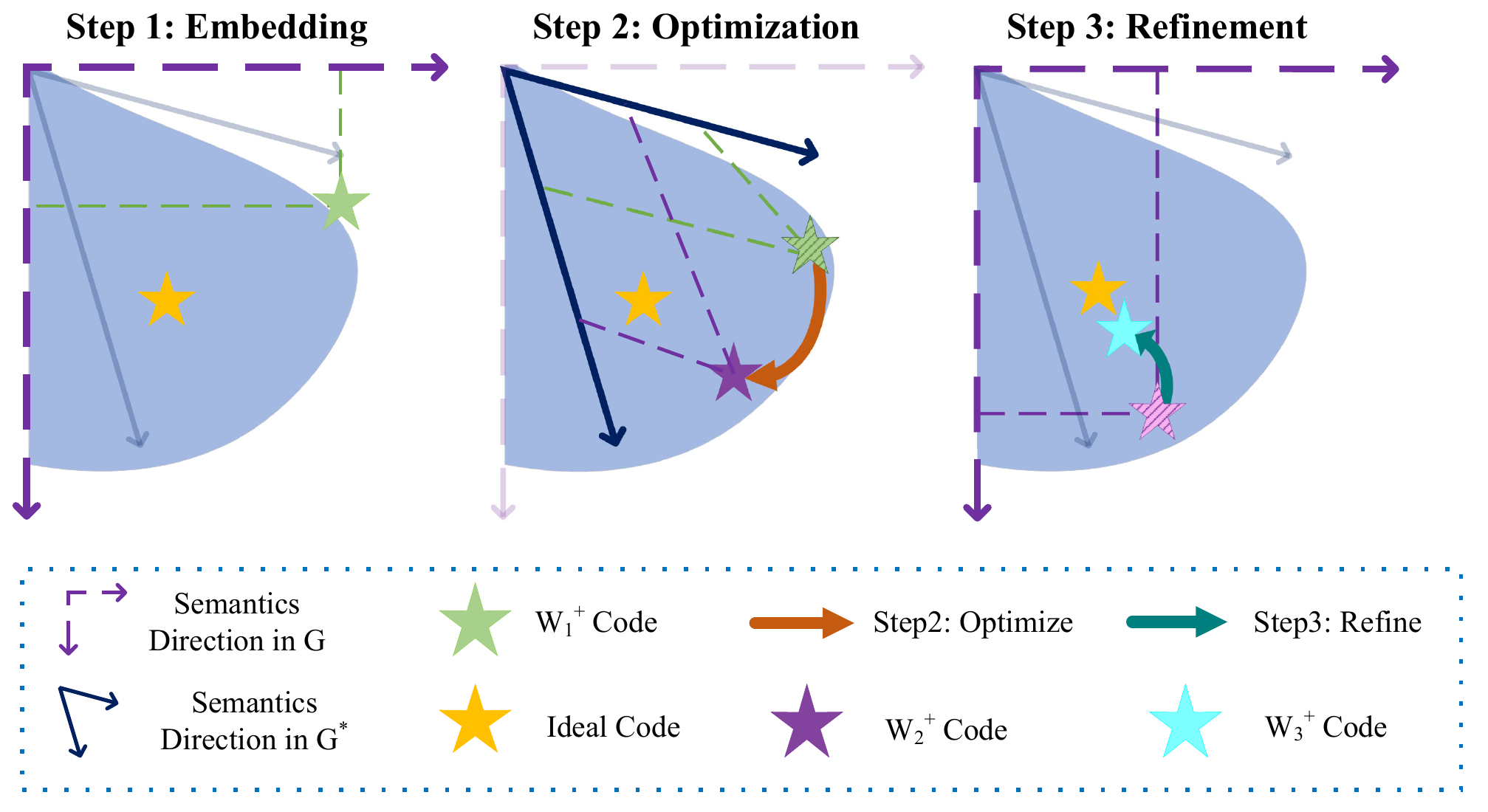}
    \caption{Pseudo-paired data generation in latent space.}
    \label{fig:multi_level_optimization}
\end{figure}

However, since semantic shifts do occur and current real-domain encoders are imperfect in content embedding, we adopt a multi-stage approach, as illustrated in Figure~\ref{fig:multi_level_optimization}, to progressively approximate the ideal image encoding, aiming to generate highly realistic and content-consistent pseudo-real portraits.

\begin{figure}[htbp]
\centering
\includegraphics[width=0.6\textwidth]{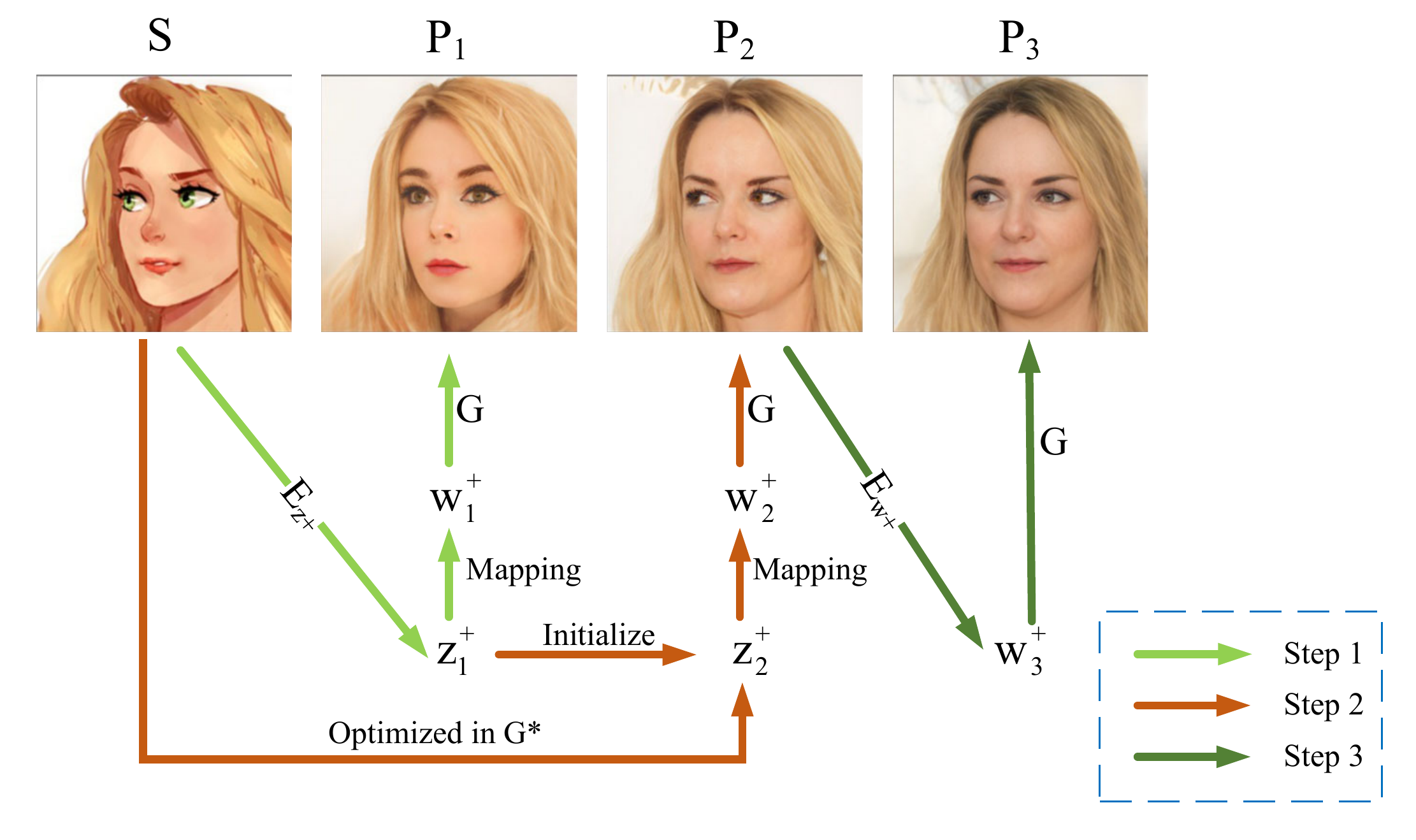} 
\caption{Pseudo-paired data generation process: $G^*$ denotes a StyleGAN finetuned without proposed constraints, and \textit{Mapping} refers to the mapping network of $G$. $E_{z^+}$ and $E_{w^+}$ donates encoder mapping images into $Z^+$ and $W^+$ space, respectively.} 
\label{Makedata}
\end{figure}

StyleGAN primarily utilizes two latent spaces, $Z$ (or $Z^+$) and $W$ (or $W^+$). The $Z^+$ space follows an extended normal distribution, and with appropriate truncation tricks, encodings can be mapped onto the distribution center to generate portraits of high visual quality. It serves as a latent space for \textit{embedding} and \textit{optimization}. In contrast, the $W^+$ space is semantically rich for real portraits, capable of adding detailed and realistic touches, making it suitable for \textit{refinement}. The process of generating pseudo-paired data is illustrated in Figure \ref{Makedata}.

\noindent \textbf{Embedding} We adjust a pSp encoder pretrained on the FFHQ dataset to map the style image  \( S \) into the $Z^+$ space of generator $G$. This yields the first-level realistic image \( P_1 \) and its corresponding encodings \( z^+_1 \) and \( w^+_1 \).
\begin{equation}
z^+_1=E_{z^+}\left(S\right),w^+_1=G_{mapping}\left(z^+_1\right),P_1=G\left(w^+_1\right)
\end{equation}

\noindent \textbf{Optimization} To enhance realism and  content consistency, \( w^+_1 \) is optimized to approach the ideal code of \( S \). Assuming \( S \) can be expressed through \( G^* \), we begin with \( z^+_1 \) and apply semantic constraint to find \( z^+_2 \) and \( w^+_2 \), generating a more realistic image \( P_2 \). 
\begin{equation}
\begin{aligned}
z^+_{2} =& \arg\min \mathcal{L}_{\mathrm{semantic}}\left(G^*(z^+), S\right) \\ 
w^+_{2} =& G_{\text{mapping}}(z^+_{2}), P_{2} = G(w^+_{2})
\end{aligned}
\end{equation}

\noindent \textbf{Refinement}
Due to semantic discrepancies between the two models, $P_2$ may still lack authenticity and fidelity. Therefore, we refine $P_2$ using a pSp encoder pretrained in $W^+$ space to preserve image details and enhance realism.
\begin{equation}
w^+_3=E_{w^+}\left(S\right),\quad P_3=G\left(w^+_3\right)
\end{equation}

\begin{figure}[htbp]
\centering
\includegraphics[width=0.6\textwidth]{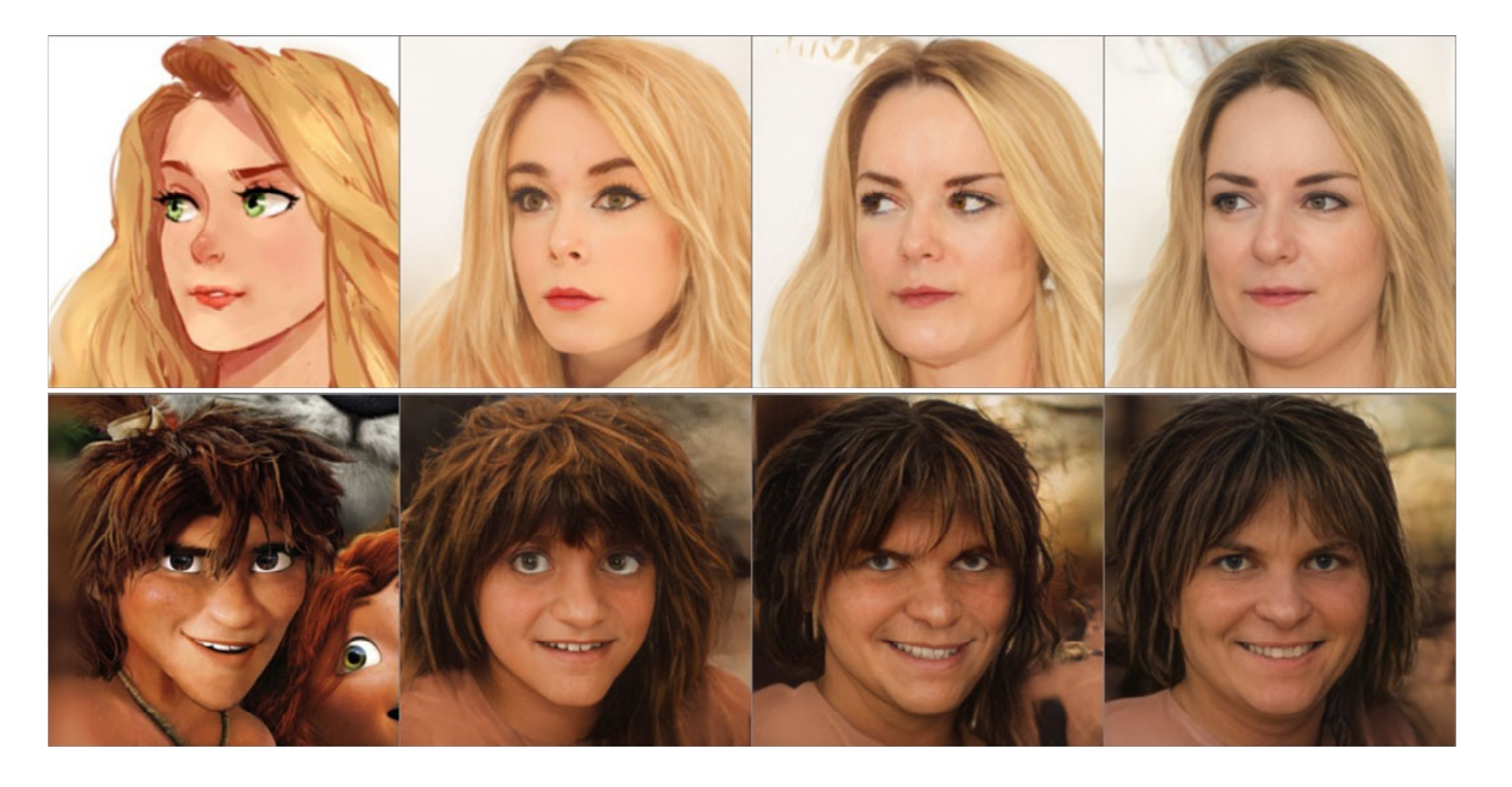} 
\caption{Example of pseudo-paired data} 
\label{paired_example}
\end{figure}
Through the aforementioned method, we obtain three levels of realistic-stylized portrait paired data as in Figure \ref{paired_example}, each with corresponding latents. Depending on the characteristics of different styles, the appropriate level of paired data can be selected.
\subsection{Multimodal and Reference-Guided Stylization}
Our method supports multimodal and reference-guided portrait stylization by encoding content in latent space and blending at different scales with random noises or embeddings of reference images, achieving varied stylization without additional network structures or training.

We find that accurate embedding into the latent space is critical for effective stylization. Based on this insight, we made the following adjustments:

For multimodal portrait stylization: We sample from $Z^+$ space to introduce diverse noise, combining it with truncation tricks to map it onto concentrated regions of $W+$ space, ensuring high-quality portrait generation.

For reference-guided portrait stylization: We use the GAN inversion method from \cite{UI2I-style} to optimize the embedding of the reference image into the $V$ space of generator $G'$, converting it to $W$ space and then replicating it into $W^+$ space, ensuring the encoding remains within the latent space.
We refer to Section 4.4 for more stylization results.

\section{Experiment}
\label{sec:experiment}

\subsection{Experimental Settings}
\noindent \textbf{Dataset}
For finetuning, our dataset comprises 317 cartoon-style images from Toonify \cite{toonify}, 174 anime-style images from Danbooru \cite{danbooru2019Portraits}, 137 fantasy-style, 156 illustration-style, and 120 impasto-style images from \cite{dualstylegan}. For testing, facial images from the FFHQ dataset \cite{stylegan} are utilized. All images are resized to a resolution of 1024.

\noindent \textbf{Compared Methods}
Our method is compared with mainstream portrait stylization approaches: I2I method U-gat-it \cite{ugatit}, StyleGAN-based method Toonify \cite{toonify}, UI2I-style \cite{UI2I-style}, DualStyleGAN \cite{dualstylegan}, and diffusion-based method NTC \cite{ntc}, InST \cite{inst}, and StyleID \cite{styleid}.

\noindent \textbf{Evaluation Metrics}
Portrait stylization is evaluated based on stylization effect (quality) and content consistency (fidelity). Objectively, we utilize the Fréchet Inception Distance (FID)\cite{FID} to measure stylization effect and perceptual loss\cite{perceptual-loss} to assess fidelity. Subjectively, we conduct a user survey involving 50 volunteers who rate the stylized results on a scale from 0 to 5 in terms of quality and fidelity. (For each style, five randomly selected outputs are evaluated.)

Additional details regarding the experimental setup are provided in the supplementary material.

\subsection{Comparative Experiments}
\begin{figure*}[htbp]
\centering
\includegraphics[width=1\textwidth]{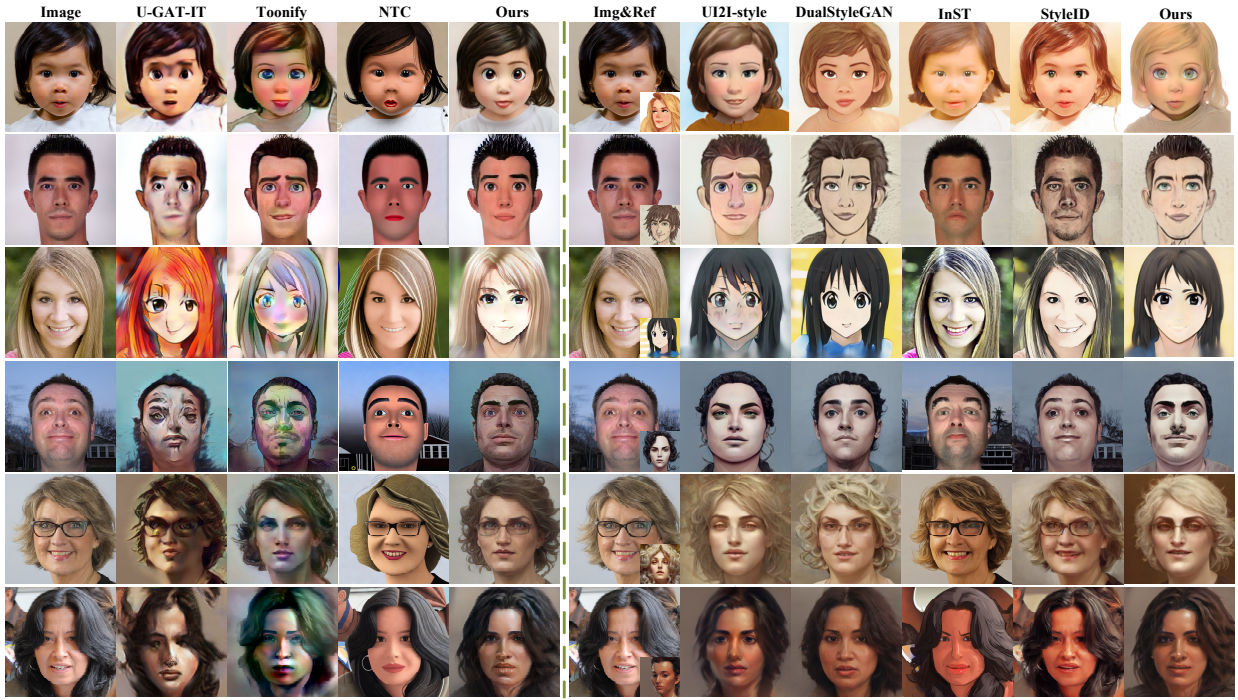} 
\caption{Qualitative results of the comparative experiment, styles from top to bottom: Cartoon, Anime, Fantasy, Illustration, Impasto.} 
\label{compare_exp}
\end{figure*}

Figure \ref{compare_exp} presents qualitative comparison results. U-gat-it suffers from inaccurate facial structures due to complex mappings learned from scratch. For StyleGAN-based methods, Toonify introduces artifacts, while UI2I-style and DualStyleGAN fail to maintain content consistency with input portraits in referenced stylization. These issues arise from the lack of constraints on the semantic shift of the generator, resulting in poor stylization quality and low fidelity. In contrast, our method achieves better content consistency \textit{(such as hairstyle, facial contours, and facial features)} while maintaining sufficient stylization effects.

Compared with diffusion-based methods, NTC achieves stylization mainly by blurring and simplifying textures. However, this approach is unsuitable for highly abstract styles like anime and cartoons or complex-textured styles like fantasy. Also, NTC's sampling process introduces perturbations, leading to results that mismatch the original image, such as the radial hair in the third row and the earrings in the last row. InST and StyleID only transfer low-level visual features like color and texture, failing to achieve higher-level abstractions, such as geometric deformations or facial feature changes. This supports our hypothesis that current diffusion-based methods lack the ability to extract high-level style semantics from reference images without direct text guidance.


Quantitative results are shown in Tables \ref{tab:comparative results} and \ref{tab:user survey}. Our method outperforms GAN-based approaches.  While diffusion-based methods perform better in fidelity quantitative metrics,  we still consider our method more effective because diffusion-based methods do not exhibit sufficient stylization effects (e.g., texture simplification and geometric deformations of facial features).

\begin{table}[htbp]
\centering
\begin{adjustbox}{width=0.6\columnwidth}
\begin{tabular}{ccccccc}
\hline
\multirow{2}{*}{Method} & \multicolumn{2}{c}{Cartoon} & \multicolumn{2}{c}{Illustration} & \multicolumn{2}{c}{Anime} \\ \cline{2-7} 
                        & FID↓ & Perc.↓ & FID↓ & Perc.↓ & FID↓ & Perc.↓ \\ \hline
U-gat-it \cite{ugatit}   & 153.94 & 0.465 & 119.20 & 0.543 & 146.12 & 0.583 \\
Toonify \cite{toonify}    & 166.57 & 0.478 & 97.02 & 0.541 & 135.43 & 0.587 \\
NTC \cite{ntc}       & 166.04 & \bfseries  0.403 & 164.66 & \bfseries 0.403 & 198.70 & \bfseries 0.403 \\
Ours       & \bfseries 150.24 & \textit{0.413} & \bfseries 56.17 & \textit{0.491} & \bfseries 99.13 & \textit{0.559} \\ \hline
UI2I-style \cite{UI2I-style} & 235.14 & 0.561 & 133.72 & 0.559 & 142.25 & 0.653 \\
DualStyleGAN \cite{dualstylegan}& 276.68 & 0.547 & 127.96 & 0.509 & 164.30 & 0.665 \\
InST \cite{inst}       &211.22 &0.499 &193.08 &0.445 &257.01 &0.439 \\
StyleID \cite{styleid}    &224.00 & \bfseries 0.461 &151.29 & \bfseries 0.386 &184.55 & \bfseries 0.433 \\
Ours (ref) & \bfseries 191.63 &  \textit{0.526} & \bfseries 123.72 & \textit{0.504} & \bfseries 132.09 &  \textit{0.617}    \\ \hline
\end{tabular}
\end{adjustbox}
\caption{Quantitative results of comparative experiment} 
\label{tab:comparative results} 
\end{table}

\begin{table}[htbp]
\centering
\begin{adjustbox}{width=0.6\columnwidth}
\begin{tabular}{ccccccc}
\hline
\multirow{2}{*}{Method} & \multicolumn{2}{c}{Cartoon} & \multicolumn{2}{c}{Illustration} & \multicolumn{2}{c}{Anime} \\ \cline{2-7} 
                        & Quality↑ & Fidelity↑ & Quality↑ & Fidelity↑ & Quality↑ & Fidelity↑ \\ \hline
U-gat-it \cite{ugatit}                & 1.2 & 2.3 & 1.1 & 1.5 & 1.1 & 1.8     \\
Toonify \cite{toonify}               & 3.5 & 3.7 & 2.7 & 2.8 & 2.9 & 3.2     \\
NTC \cite{ntc}                    & 2.8     & 3.8      & 3.5     & 3.8      & 2.5     & \bfseries 4.0     \\
 Ours    & \bfseries 4.5      & \bfseries 4.0 & \bfseries 4.3 & \bfseries 3.9 & \bfseries 3.9 &  3.9     \\ \hline
UI2I-style \cite{UI2I-style}     & 4.1 & 3.4 & 4.0 & 3.2 & 3.5 & 2.9     \\
DualStyleGAN \cite{dualstylegan}          & 4.4 & 3.8 & 4.2 & 3.5 & 4.0 & 3.8      \\
InST \cite{inst}                   & 3.0    &4.0       & 3.5     & 3.7       & 2.3     & \bfseries 4.1     \\
StyleID \cite{styleid}                & 3.2    & \bfseries 4.3       &  3.6     & \bfseries 4.0       & 2.2     & 4.0      \\ 
Ours (ref)    & \bfseries 4.5 & 4.1 & \bfseries 4.4 &  3.8 & \bfseries 4.1 &  4.0    \\ \hline
\end{tabular}
\end{adjustbox}
\caption{User survey results} 
\label{tab:user survey} 
\end{table}
Additionally, we evaluated model sizes, image resolutions, train and test time in Table \ref{tab:model_size}. Our method can achieve high-quality stylization with relatively fewer computational resources compared with the GAN-based approach, While the diffusion-based method does not require training, its real-time performance is limited by module size and longer inference time.

\begin{table}[htbp]
\centering
\begin{adjustbox}{width=0.6\columnwidth}
\begin{tabular}{ccccc}
\toprule
Model & Params (M) & Res. & Test Time (s) & Train Time (h) \\
\midrule
Toonify \cite{toonify}    & 28.27   & 1024    & 94.0   & 0.5   \\
UI2I-style \cite{UI2I-style} & 28.27   & 1024    & 96.0   & 0.5   \\
DualStyleGAN \cite{dualgan} & 354.46  & 1024    & 0.4    & 20.2  \\
NTC \cite{ntc}         & 865.70   & 512    & 5.2   & 0   \\
InST \cite{inst}       & 865.70  & 512    & 5.1    & 0  \\
StyleID \cite{styleid}    & 865.70  & 512    & 4.0    & 0  \\
Ours       & 87.00   & 1024    & 0.09   & 0.5   \\
\bottomrule
\end{tabular}
\end{adjustbox}
\caption{Model evaluation: size, resolution, and train/test time}
\label{tab:model_size}
\end{table}
\subsection{Ablation Study}
\begin{figure}[!htbp]
\centering
\includegraphics[width=.7\textwidth]{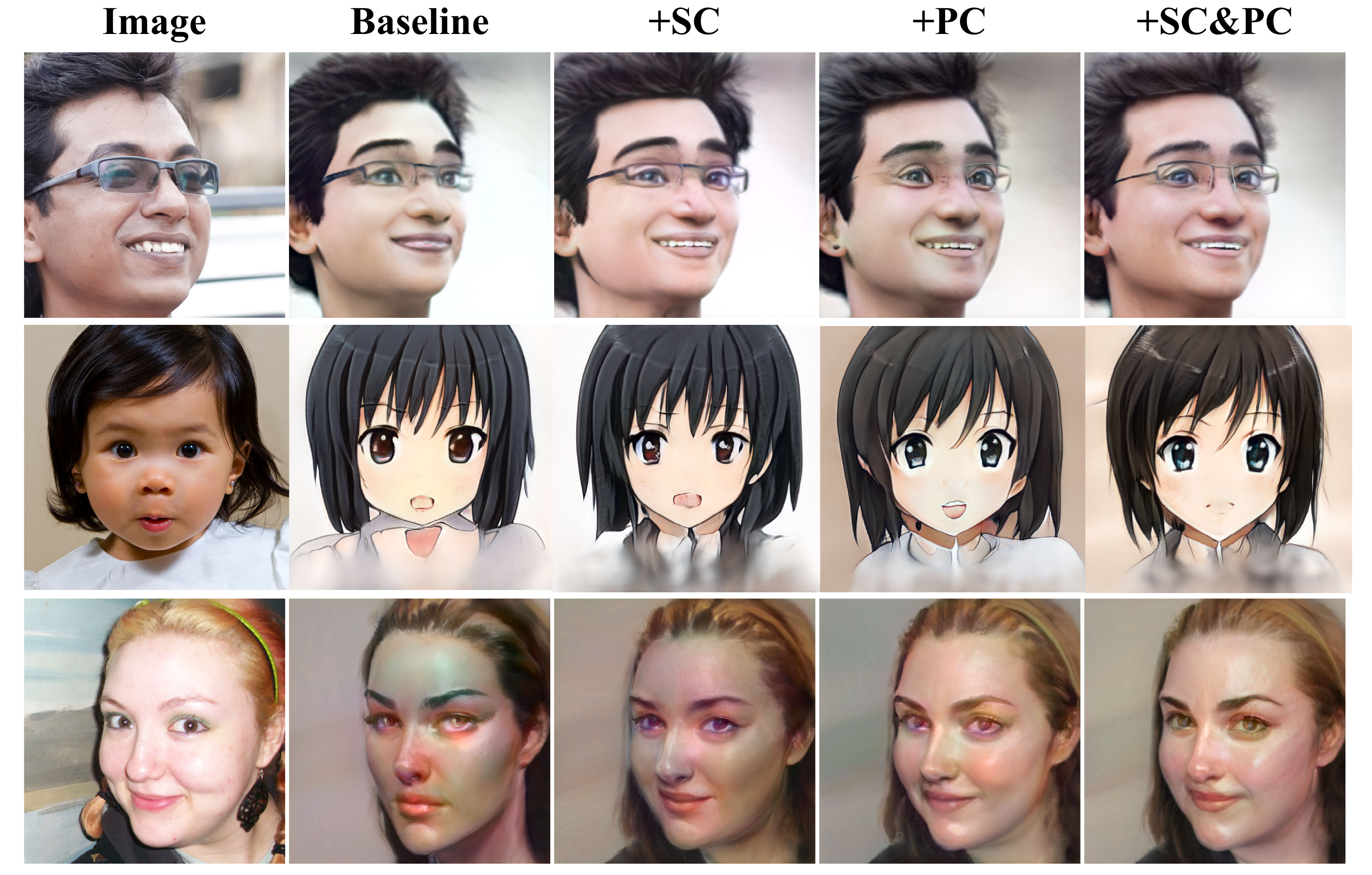} 
\caption{Ablation study of semantic and pseudo-paired constraints. SC represents semantic constraint, and PC represents pseudo-paired supervision.} 
\label{ablation-sc}
\end{figure}

We investigated the effectiveness of the proposed semantic constraint and pseudo-paired supervision. As shown in Figure \ref{ablation-sc}, both constraints significantly enhance the quality of stylization and strengthen the content correlation between input and output images. This improvement is due to the constraints limiting semantic shift during finetuning, aligning the semantics of the finetuned model more closely with those of the pretrained model. Consequently, the finetuned model inherits the rich content diversity of the pretrained StyleGAN and ensures a more consistent expression of the same latent across different domains.

To further quantify the effects of proposed constraints, we computed the semantic distance \cite{UI2I-style} between $G$ and $G'$, as well as the FID score to corresponding dataset. As shown in Table \ref{tab:ablation-sc}, the results not only demonstrate that our improvements enhance the stylization effect but also indicate that the proposed constraints make the fine-tuned models semantically closer to the pre-trained model.

\begin{table}[htbp]
\centering
\begin{adjustbox}{width=0.6\columnwidth}
\begin{tabular}{ccccccccc}
\hline
\multirow{2}{*}{Dataset} & \multicolumn{2}{c}{Baseline} & \multicolumn{2}{c}{+SC} & \multicolumn{2}{c}{+PC} & \multicolumn{2}{c}{+SC\&PC} \\
\cline{2-9} 
 & FID & Dis. & FID & Dis. & FID & Dis. & FID & Dis. \\
\hline
Cartoon        & 169.54 & 0.570 & 153.72 & 0.492 & 151.65 & 0.477 & \bfseries 150.24 & \bfseries 0.413 \\
Anime          & 149.63 & 0.648 & 134.33 & 0.603 & 115.62 & 0.582 & \bfseries 99.13 & \bfseries 0.559 \\
Fantasy        & 149.57 & 0.613 & 138.52 & 0.586 & 118.68 & 0.537 & \bfseries 110.25 & \bfseries 0.489 \\
Illustration   & 115.64 & 0.617 & 98.17 & 0.591 & 94.57 & 0.540 & \bfseries 56.17 & \bfseries 0.491 \\
Impasto        & 134.19 & 0.549 & 123.51 & 0.513 & 106.12 & 0.484 & \bfseries 102.95 & \bfseries 0.428 \\
\hline
\end{tabular}
\end{adjustbox}
\caption{Quantitative evaluation of ablation study on constraints.}
\label{tab:ablation-sc}
\end{table}

Additionally, we find that different style categories are suitable for encoding at varying levels of pseudo-paired data (see Section 4 of the supplementary material). We also investigate the impact of content embeddings in different latent spaces on portrait stylization (detailed in Section 2 of the supplementary material).

\subsection{Multimodal and Reference-Guided Stylization}
\begin{figure}[!htbp]
\centering
\includegraphics[width=.7\textwidth]{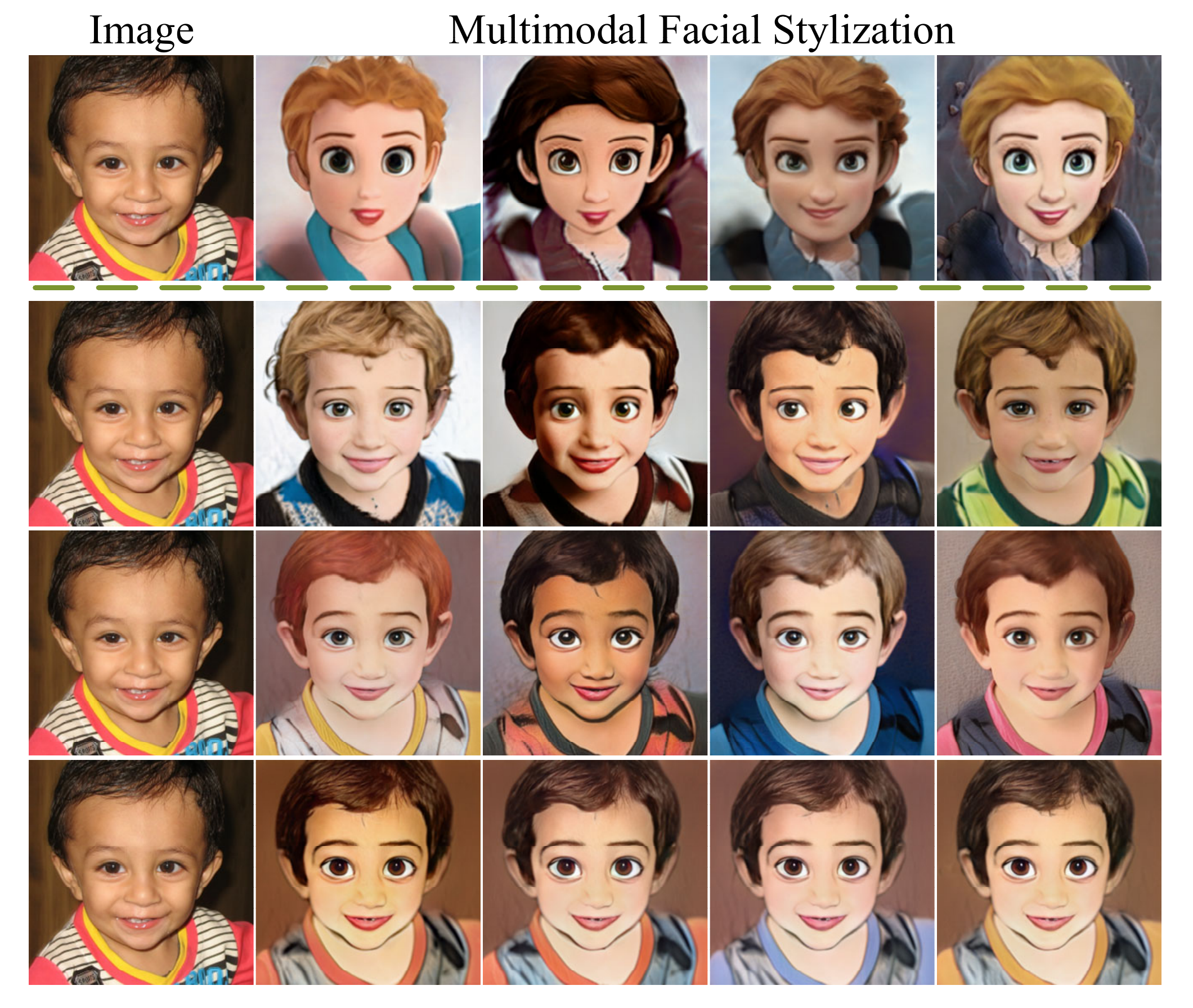} 
\caption{Multimodal stylization results. First row: UI2I-style method \cite{UI2I-style}. Second row and below: Our method with encoding combinations 6, 9, and 12.} 
\label{multimodal}
\end{figure}
Figure \ref{multimodal} illustrates the results in multimodal portrait stylization. While the UI2I-style method generates diverse portraits, it lacks practical semantic constraints, resulting in lower-quality outputs. Our method achieves controlled diversification through style mixing at various levels. low mixing layers preserve essential facial characteristics, with changes primarily affecting hairstyles and attire. As mixing level increases, variations become more subtle, influencing attributes such as hair color, skin tone, and clothing color.
\begin{figure}[htbp]
\centering
\includegraphics[width=.7\textwidth]{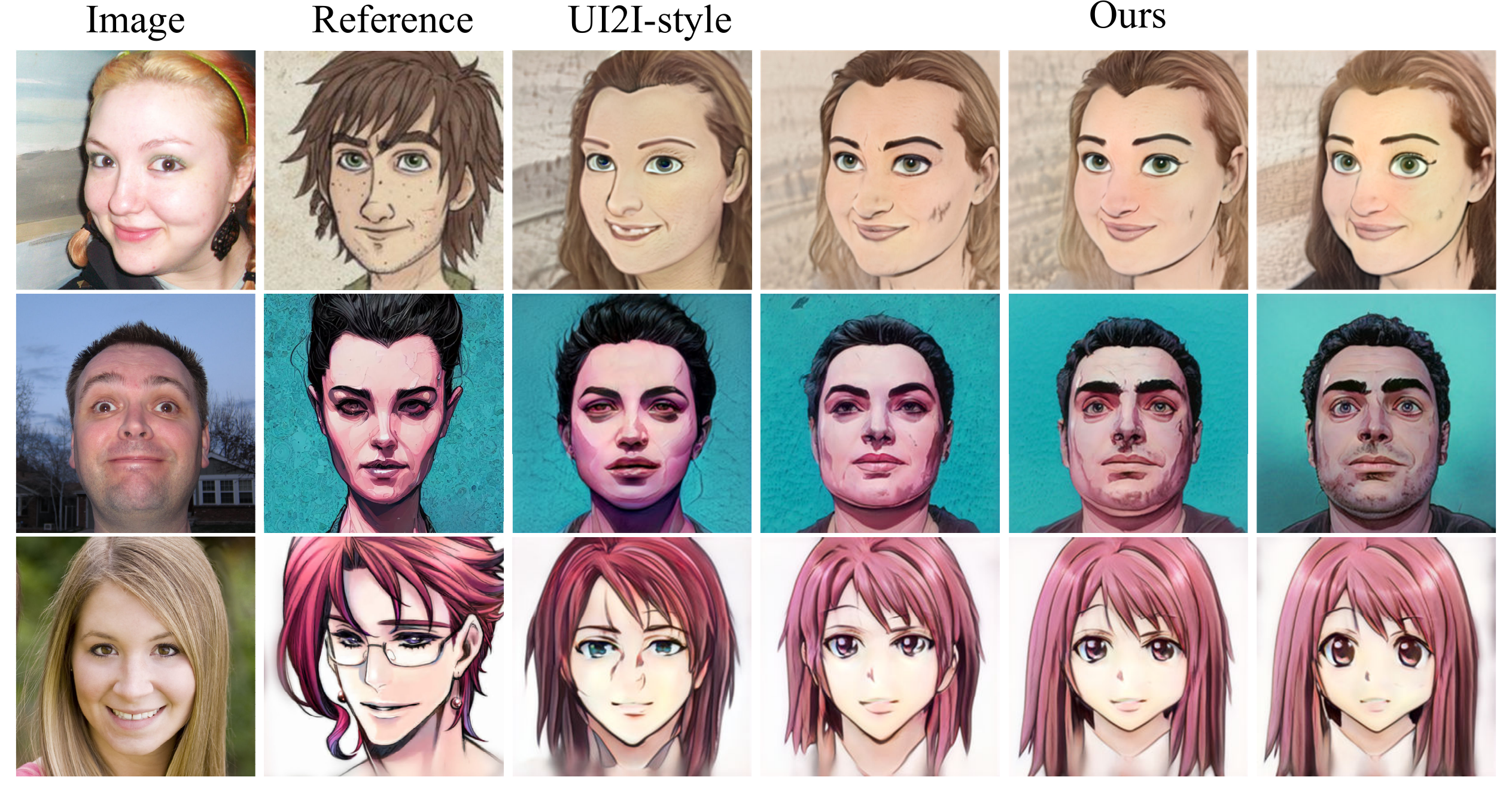} 
\caption{Reference-guided stylization results. Mixed encoding combinations: 3, 6, and 9.} 
\label{ref}
\end{figure}

As Figure \ref{ref} demonstrates, our method yields high-quality results in reference-guided portrait stylization. With lower-level mixing, the generated portraits inherit more characteristics from the reference image, such as the masculine eyebrows and eyes shown in the first row. As the mixing layer increases, the generated images retain refined features from the reference, including hair and skin color.

Additionally, we investigate the impact of style encodings in various latent spaces on reference-guided portrait stylization. Details are provided in Section 3 of the supplementary material.
\subsection{Pseudo-Paired Data in Different Latent Spaces}

\begin{figure}[!htbp]
\centering
\includegraphics[width=.6\textwidth]{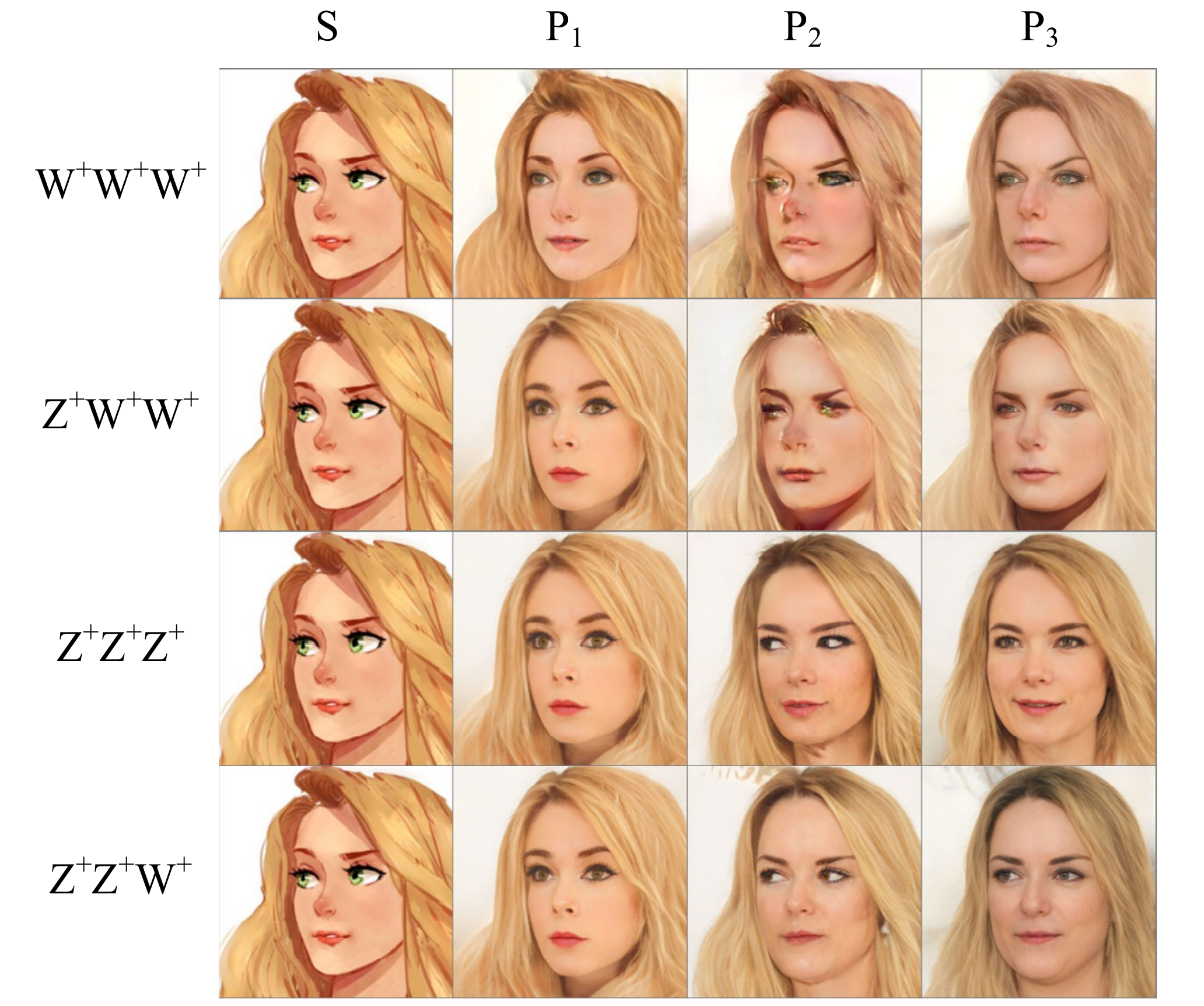} 
\caption{Paired data obtained by sequentially operating on images in different latent spaces.} 
\label{datapair}
\end{figure}

Figure \ref{datapair} shows the multi-level pseudo-paired data obtained by sequentially encoding style images in different latent spaces. Although the $W^+$ space is semantically rich, directly \textit{embedding} $S$ into this space degrades image quality and leads to poor content correspondence due to semantic shifts. Similarly, \textit{optimization} in the $W^+$ space exacerbates semantic shifts, resulting in obvious artifacts. Therefore, we initially use the $Z^+$ space to ensure image quality and subsequently leverage the rich semantic information of the $W^+$ space during \textit{refinement} to enhance realistic details and content consistency (such as the gaze direction in $P_3$).

\section{Conclusion}
\label{sec:conclusion}
In this paper, we present a facial stylization approach based on StyleGAN. By introducing semantic constraint loss and pseudo-paired supervision, we effectively mitigate semantic drift caused by changes in data distribution during finetuning, thereby achieving higher quality and more faithful stylization results. Additionally, we developed a method for generating multi-level pseudo-paired data, producing paired samples with varying degrees of realism based on given stylized portraits. Finally, we achieve flexible multimodal and reference image stylization through style mixing with sampling noises and reference image encodings at different levels. Experimental results demonstrate that our method produces more appealing and content-faithful stylized portraits than previous methods.

\section{Appendix}

\subsection{Implementations in Model Training and Testing}

\noindent \textbf{Training Details} Portrait stylization training (as well as testing) are conducted on one NVIDIA Tesla V100 GPU using PyTorch, with the Adam \cite{adam} optimizer and a learning rate of 0.02. The steps involving optimization in this paper include the generation of pseudo-paired data pairs, model fine-tuning, and encoding of reference images.

\noindent \textit{For paired data generation}, the optimization process is set to 1000 iterations with a batch size of 1 and $\lambda_{id} = 0.1$; generating a set of paired data takes approximately 1.5 minutes.

\noindent \textit{During finetuning}, the batch size is 4, $\lambda_{semantic} = \lambda_{paired} = 1$. The model typically converges within 1000 iterations, with an average training time of 0.5 hours per style. 

\noindent \textit{For embedding of reference image}, We use the method from \cite{UI2I-style}, with a batch size of 1. It takes approximately 90 seconds to obtain the embedding for each image. Once a reference image has been encoded, it can be continuously utilized in subsequent tests.

\noindent \textit{About perceptual and identity loss}, To reduce computational cost, the image resolution is adjusted to 256 when computing $\mathcal{L}_\mathrm{LPIPS}$ and $\mathcal{L}_\mathrm{ID}$. The pretrained-model of VGG \cite{vgg} is used for $\mathcal{L}_\mathrm{LPIPS}$ during paired data creation and finetuning, while that of AlexNet \cite{alexnet} is utilized for evaluating the quantitative metric Perceptual Loss.

\noindent \textbf{Encoder} We require a pSp encoder \cite{psp} that maps real facial images to the $W$ latent space for portrait stylization. Additionally, pSp encoders that map to the $W^+$and $Z^+$ spaces are needed to generate pseudo-paired data.

pSp encoder provides the $W^+$space encoder, whereas the other two encoders are derived by simply modifying its architecture and training on the FFHQ dataset for a image reconstruction task: sharing parameters across its mapping units adjusts to $W$ space, while retaining StyleGAN's mapping network allows adjustments to $Z^+$ space.

\noindent \textbf{Generator} We finetuned StyleGAN for five styles: cartoon, anime, fantasy, illustration, and impasto. According to the training strategy described in the paper, the cartoon, fantasy, illustration, and impasto styles converged after 1000 iterations. The anime style required 3000 iterations due to its greater divergence from the real domain. During testing, the truncation trick was set to 0.7 for cartoons, 0.6 for anime to minimize artifacts, and 0.9 for the other styles to enhance fidelity.

\noindent \textbf{Hyperparameter Search} Our hyperparameters primarily focus on semantic preservation loss and pseudo-paired supervision. We fix the ratio of LPIPS to identity loss at 1:0.1 and adjust \(\lambda_{\text{semantic}}\) in steps of 10 times, ranging from 0.001 to 10. We find that a weak semantic preservation loss fails to effectively maintain image quality and fidelity, while a strong one diminishes the stylization effect and introduces artifacts. The optimal \(\lambda_{\text{semantic}}\) is determined to be 1. Based on this, we introduce pseudo-paired supervision, tuning \(\lambda_{\text{paired}}\) from 0 to 5 in increments of 0.5. The best \(\lambda_{\text{paired}}\) value is set to 1. Similar to \(\lambda_{\text{semantic}}\), excessive \(\lambda_{\text{paired}}\) values result in more noticeable artifacts. In our experiments, the latent variable \(W^+_1\) is used for anime style in the comparative experiments, while \(W^+_2\) supervision is applied to all other styles; further details on the study of latent variable levels are provided in Section 4 of the supplementary material.

\subsection{Study on Content Encodings in Different Latent Spaces}

\begin{figure}[!h]
    \centering
    \includegraphics[width=.6\textwidth]{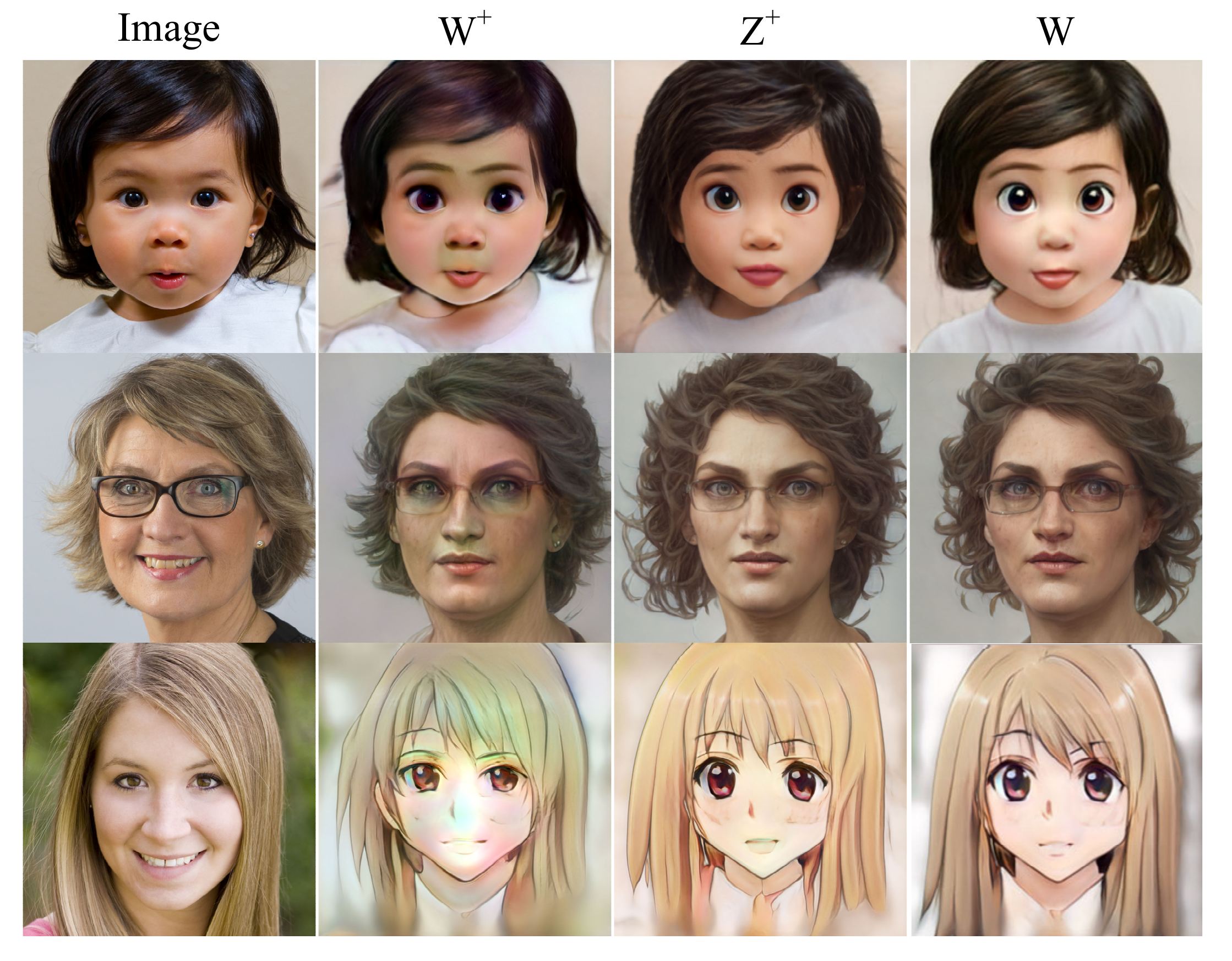}
    \caption{Stylization results of encoding the input image to different latent spaces. We use a modified pSp encoder to encode the original images into the \(W^+\), \(Z^+\), and \(W\) domains.}
    \label{sup-1}
\end{figure}

We examine how encoding content vectors in various latent spaces influences portrait stylization. As shown in Figure \ref{sup-1}, experimental findings indicate that encoding in the $W^+$ space introduces artifacts into the generated images. This could be due to the $W^+$space retaining detailed information from the real domain, which can be exaggerated during the stylization process and lead to artifacts. Conversely, encoding in the \(Z^+\) space maps latent variables close to the concentrated distribution, thus maintaining image quality while potentially sacrificing fidelity.

For portrait stylization, we believe encoding should capture content semantics accurately while avoiding excessive real-domain-specific representation. To address this, we introduce a modified pSp encoder that maps images to the \(W\) space, achieving a balance between fidelity and stylization quality. 

\subsection{Study on Style Encodings in Different Latent Spaces}
\begin{figure}[htbp]
    \centering
    \includegraphics[width=.6\textwidth]{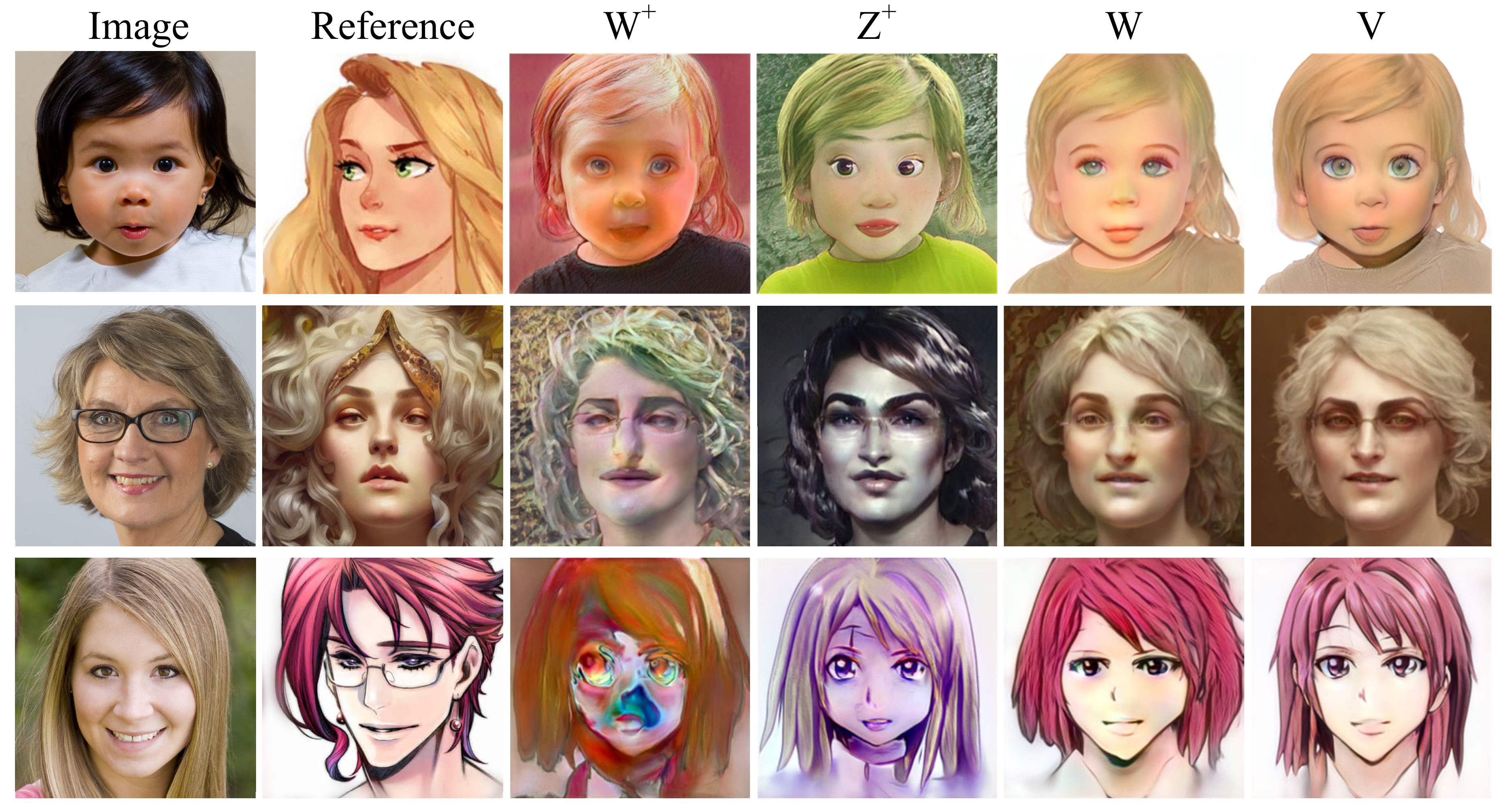}
    \caption{Stylization results of encoding the reference image to different latent spaces. We adjust \cite{stylegan2} to optimize the reference images into \(W^+\), \(Z^+\), and \(W\) spaces, and use the method in \cite{UI2I-style} first encodes the image into the \(V\) space and then maps it back to the \(W\) domain.}
    \label{sup-2}
\end{figure}

Figure \ref{sup-2} illustrates the results of encoding reference images into different latent spaces using the finetuned StyleGAN $G'$. Encoding into the $W^+$ space produces noticeable artifacts while encoding into the $Z^+$ space leads to deviations from the original content. In contrast, encoding into the $W$ space and $V$ space better preserves the style of the reference image and ensures content consistency between the generated and input images.

We propose that encoding aims to capture the style of the reference image rather than to achieve exact reconstruction. Encoding into the $W^+$ space can introduce redundancy, causing artifacts and reducing image quality. While the $Z^+$ space generates high-quality images, it captures the semantic details less accurately, leading to content discrepancies. Conversely, the $W$ and $V$ spaces perform better, with the $V$ space preserving more content information from the original image. Thus, our method favors the use of $V$ space for encoding.

\subsection{Study on Encodings for Paired Data at Different Levels in Pseudo-Supervision}

\begin{figure}[htbp]
    \centering
    \includegraphics[width=.8\textwidth]{Fig/sup-3-1.pdf}
    \caption{Results from pseudo-paired data under supervision. Pairs of rows show examples in various styles.}
    \label{df_codes}
\end{figure}

As illustrated in \ref{df_codes}, using latents from different levels of the pseudo-paired data produces varying effects. Generally, higher-level encodings represent greater discrepancies between stylized and pseudo-real images, resulting in more pronounced stylization. This observation highlights the supervisory role of paired data in managing differences between domains.

Additionally, different styles are suited to latents of varying levels. Our results show that $W^+_3$ is more effective for cartoon and impasto styles, while $W^+_1$ is preferable for anime styles. For fantasy and illustration styles, the differences across levels are less significant. A quantitative experimental result is shown in Table \ref{tab:ablation-level}. This variation could be attributed to the differing degrees of disparity between various stylized portrait and realistic face domains, affecting the pseudo-paired data at each level. We detail the multi-level pseudo-paired data for various styles in Section 5.

\begin{table}[htbp]
\centering
\begin{adjustbox}{width=0.7\columnwidth} 
\begin{tabular}{ccccccccc}
\hline
\multirow{2}{*}{Dataset} & \multicolumn{2}{c}{Baseline} & \multicolumn{2}{c}{$W^+_1$} & \multicolumn{2}{c}{$W^+_2$} & \multicolumn{2}{c}{$W^+_3$} \\
\cline{2-9} 
 & FID & Dis. & FID & Dis. & FID & Dis. & FID & Dis. \\
\hline
Cartoon        & 169.544 & 0.5705 & 152.475 & 0.4210 & 150.242 & \bfseries 0.3777 & \bfseries 148.360 & 0.4139 \\
Anime          & 149.630 & 0.6486 & \bfseries 91.626 & \bfseries 0.5092 & 99.134 & 0.5594 & 111.824 & 0.6250 \\
Fantasy        & 149.570 & 0.6135 & \bfseries 102.412 & 0.4812 & 110.246 & \bfseries 0.4888 & 133.444 & 0.4939 \\
Illustration   & 115.636 & 0.6169 & 64.263 & \bfseries 0.4615 & \bfseries 56.147 & 0.4912 & 87.802 & 0.5566 \\
Impasto        & 134.187 & 0.5487 & \bfseries 79.963 & 0.4739 & 102.952 & \bfseries 0.4279 & 113.465 & 0.4637 \\
\hline
\end{tabular}
\end{adjustbox}
\caption{Quantitative evaluation of paired data at different levels.}
\label{tab:ablation-level}
\end{table}

In summary, lower-level latents yield stable results and retain real-domain characteristics but may lack stylization. Higher-level supervision enhances stylization but risks reducing fidelity or introducing artifacts. The choice of supervision level should be based on the specific style. We suggest visualizing the pseudo-paired data before selecting the latent variable level based on discrepancies between pseudo-real and their style portrait data.

\subsection{Additional Objective Evaluation for Comparative Experiment}

To further objectively evaluate the stylization effects and fidelity, in addition to FID \cite{FID} and perceptual loss \cite{perceptual-loss} discussed in the paper, we also employed the CMMD (CLIP Maximum Mean Discrepancy) \cite{cmmd} and identity distance \cite{arcface} to measure stylization effects and fidelity, respectively. Similar to the evaluation metrics used in the paper, these two metrics leverage reliable pre-trained models to assess semantics. The results are shown in Table \ref{table1}.

\begin{table}[htbp!]
\centering
\begin{adjustbox}{width=0.7\columnwidth} 
\begin{tabular}{ccccccc}
\hline
\multirow{2}{*}{Method} & \multicolumn{3}{c}{ID Distance↓} & \multicolumn{3}{c}{CMMD↓}       \\ \cline{2-7} 
                        & Anime             & Illustration          & Cartoon           & Anime             & Illustration          & Cartoon \\ \hline
Toonify \cite{toonify}         & 0.909             & 0.781                 & 0.846             & 2.89              & 2.92                  & 2.55    \\
NTC \cite{ntc}            & \textbf{0.452}    & \textbf{0.452}        & \textbf{0.452}    & 3.29              & 4.49                  & 2.66    \\
Ours                    & \textit{\textbf{0.853}}  & \textbf{\textit{0.768}}   & \textbf{\textit{0.728}}             & \textbf{2.76}              & \textbf{2.54}                  & \textbf{1.78}    \\ \hline
UI2I-style \cite{UI2I-style}      & 0.958             & 0.846                 & 0.787             & 3.58              & 3.82                  & 3.38    \\
DualStyleGAN \cite{dualstylegan}    & 0.961             & 0.839                 & 0.790             & 3.63              & 3.59                  & 2.99    \\
InST \cite{inst}          & 0.400             & 0.512                 & 0.481             & 2.76              & 2.63                  & 2.91    \\
StyleID \cite{styleid}        & \textbf{0.253}             & \textbf{0.157}                 & \textbf{0.180}             & 3.05              & 4.48                  & 2.94    \\
Ours(ref)                    & \textbf{\textit{0.906}}             & \textbf{\textit{0.758}}                 & \textbf{\textit{0.721}}             & \textbf{2.38}              & \textbf{2.33}                  & \textbf{2.15}    \\ \hline
\end{tabular}
\end{adjustbox}
\caption{Semantic metrics between compared methods}
\label{table1}
\end{table}

From the results, it can be observed that our method achieves the best performance in terms of stylization effects and surpasses all GAN-based methods in fidelity. In contrast, diffusion-based methods exhibit only slight stylization effects (as seen from the qualitative experiments in the paper), with outputs that are almost identical to the original input images, showing changes mainly in color and low-level textures. Consequently, both identity distance and perceptual loss are lower for diffusion-based methods.

\subsection{More Results}
\begin{figure*}[p]
    \centering
    \includegraphics[width=\textwidth]{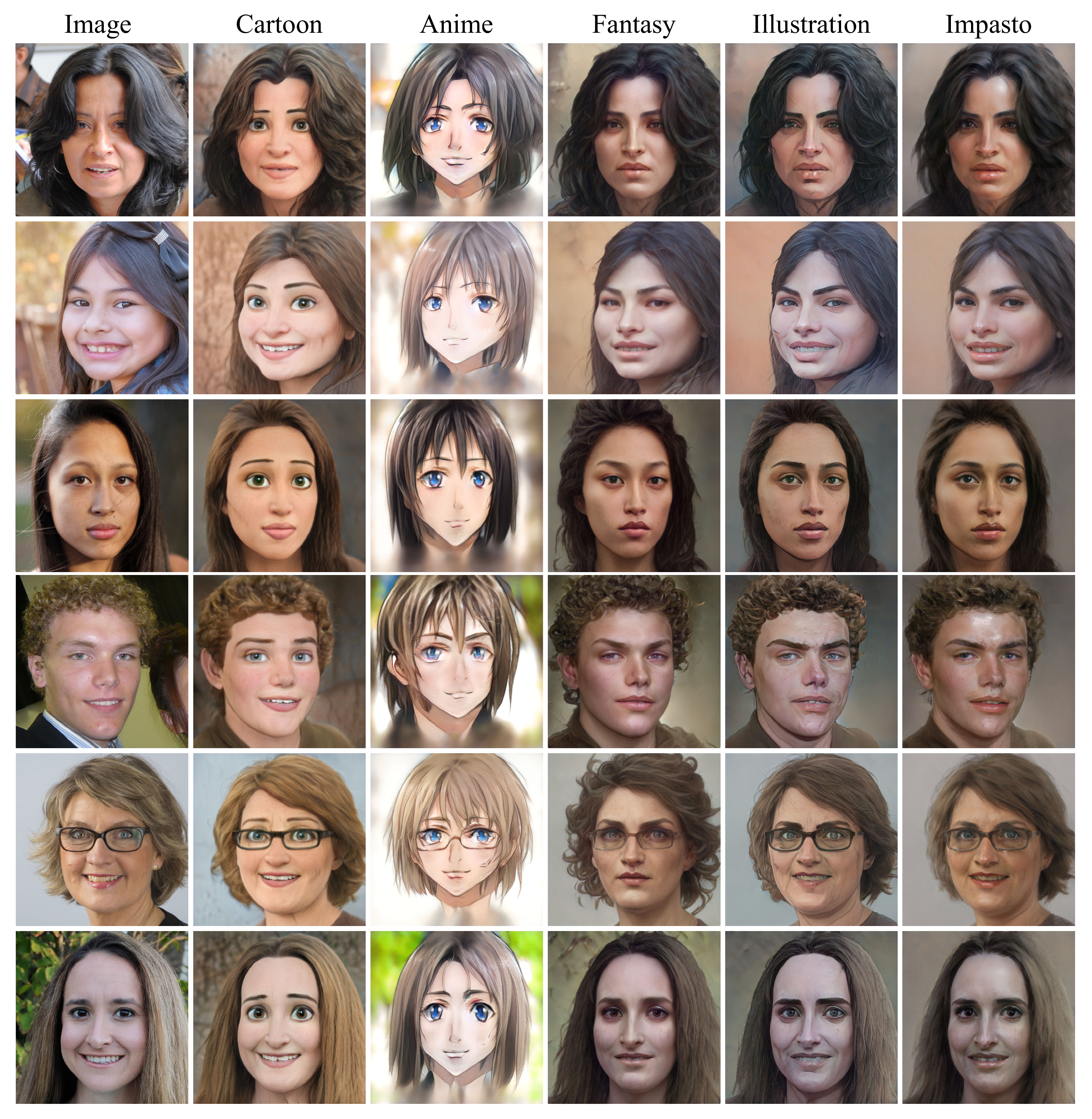}
    \caption{Portrait stylization results}
    \label{normal-1}
\end{figure*}

\begin{figure*}[htbp]
    \centering
    \includegraphics[width=\textwidth]{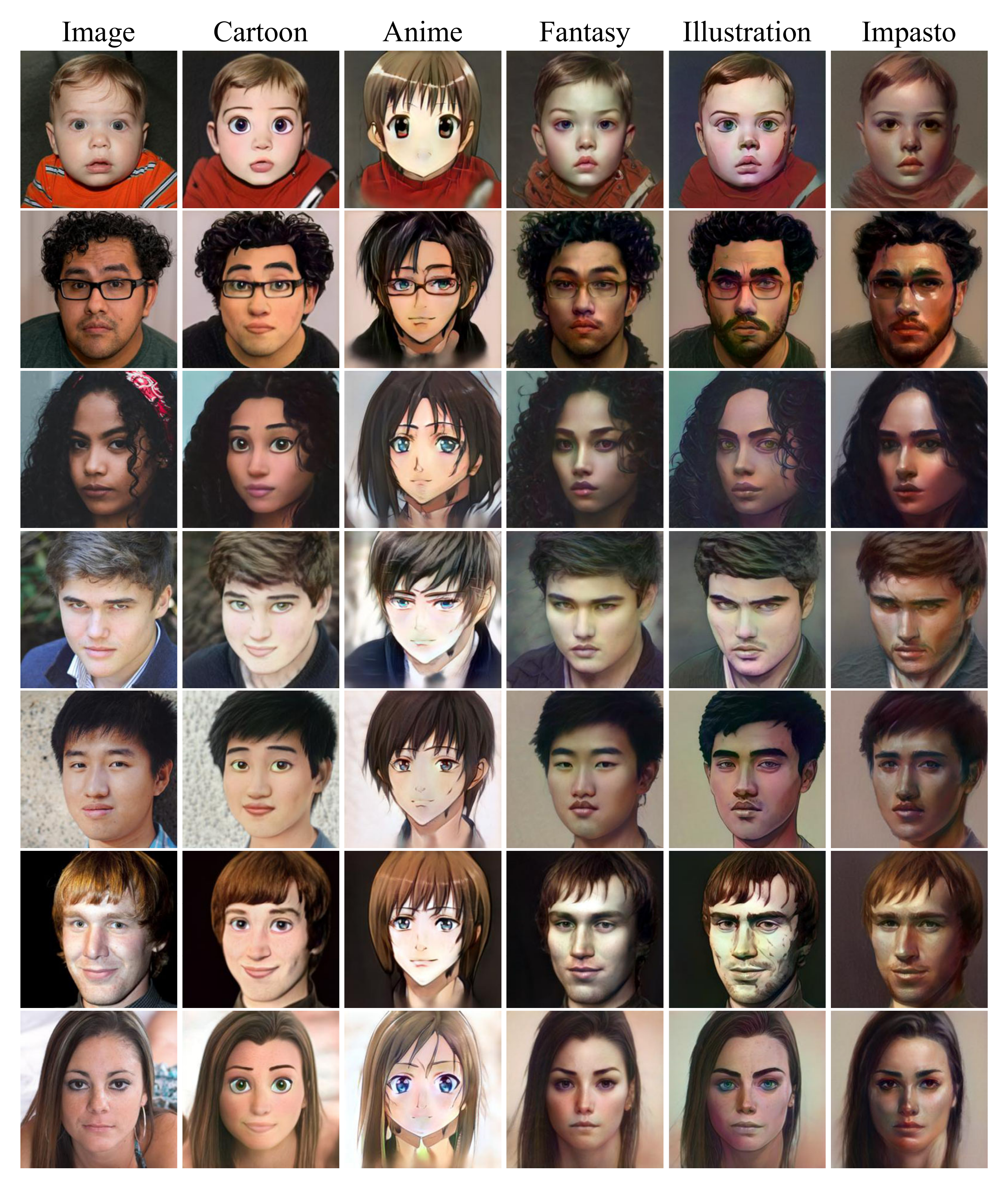}
    \caption{Portrait stylization results}
    \label{normal-2}
\end{figure*}

\begin{figure*}[htbp]
    \centering
    \includegraphics[height=\textheight]{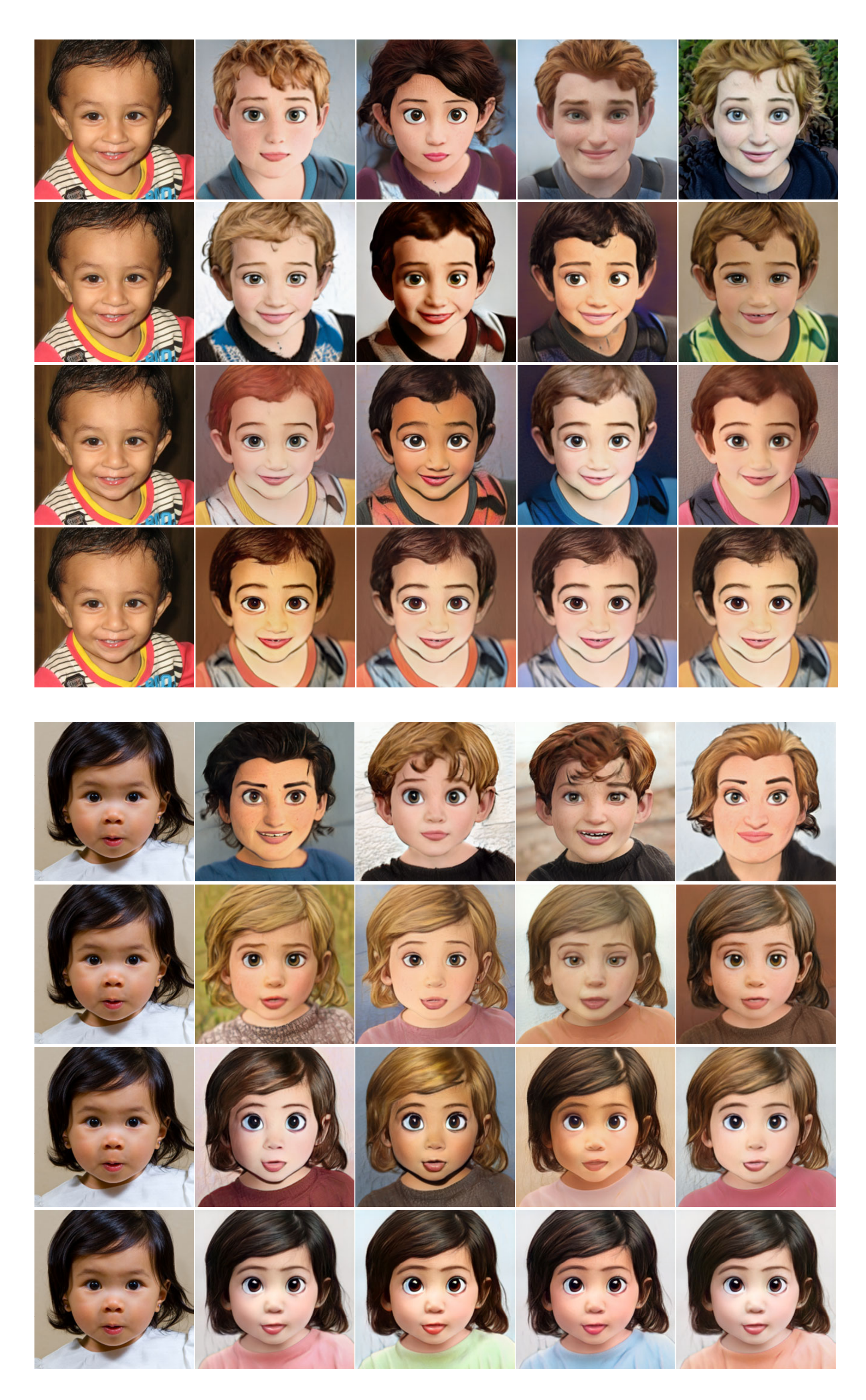}
    \caption{Multimodal portrait stylization results in cartoon style, with encoding combinations 3, 6, 9 and 12.}
    \label{c-m}
\end{figure*}

\begin{figure*}[htbp]
    \centering
    \includegraphics[height=\textheight]{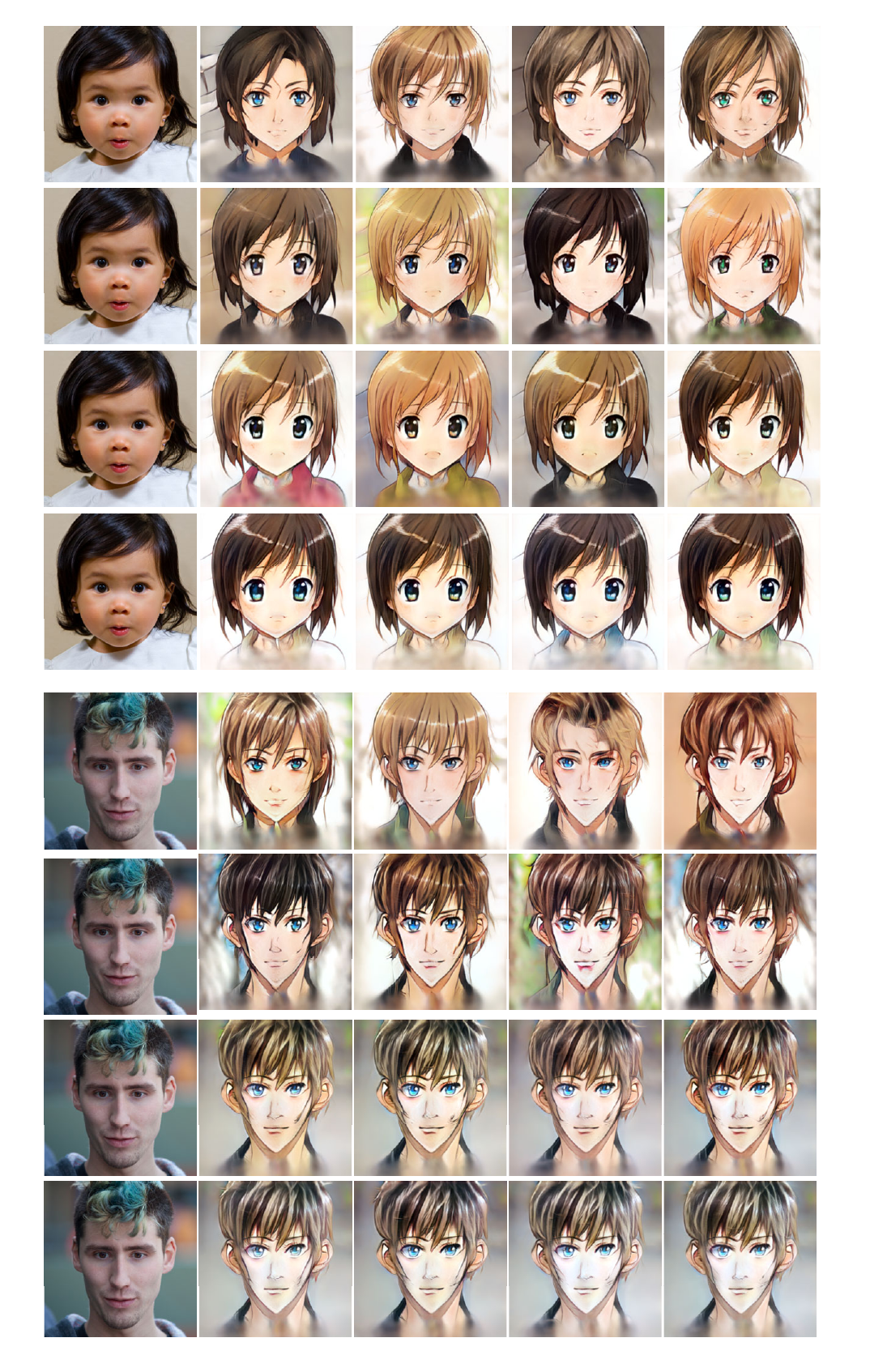}
    \caption{Multimodal portrait stylization results in anime style.}
    \label{anime-m}
\end{figure*}

\begin{figure*}[htbp]
    \centering
    \includegraphics[height=\textheight]{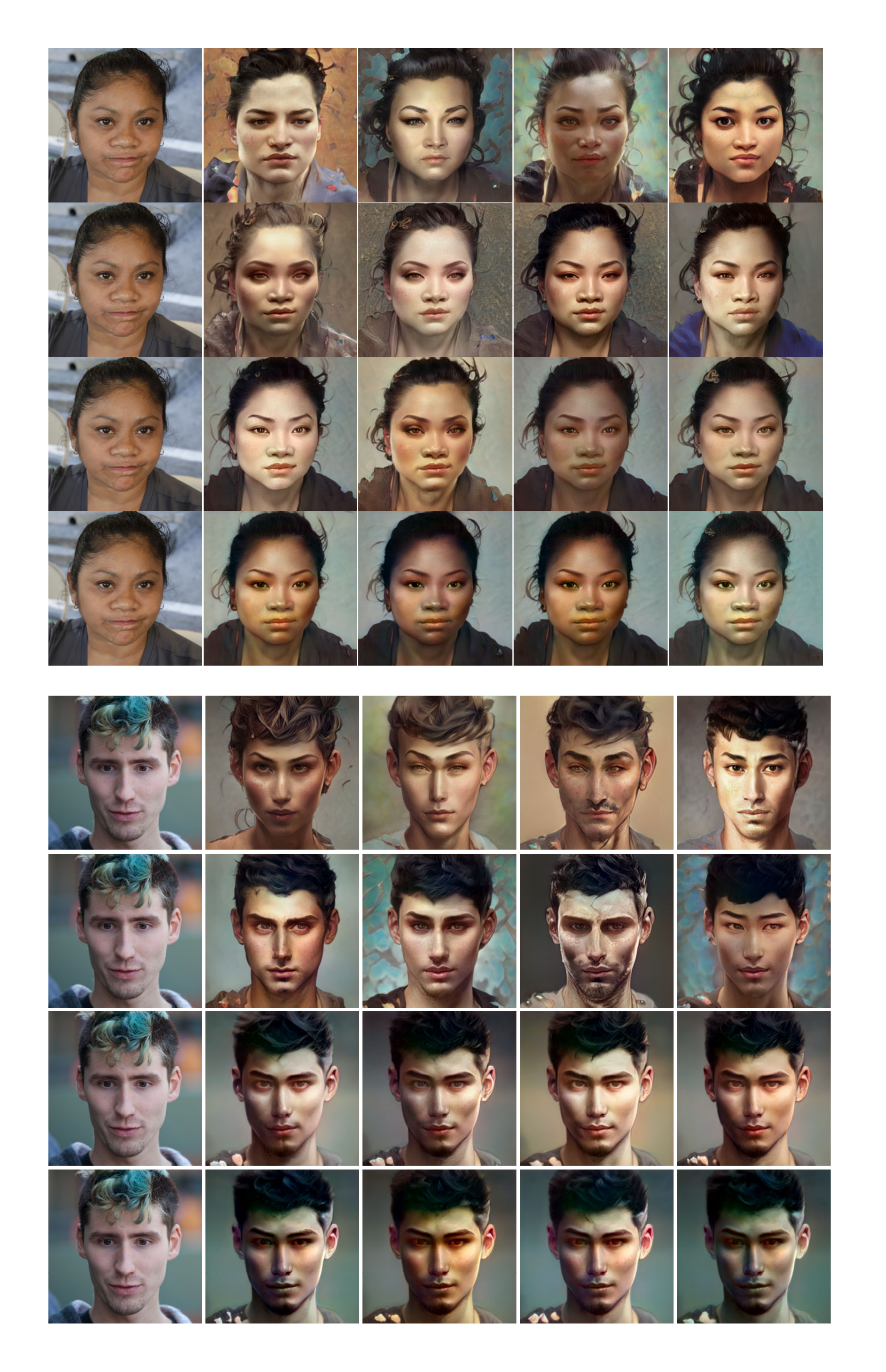}
    \caption{Multimodal portrait stylization results in fantasy style.}
    \label{fantasy-m}
\end{figure*}

\begin{figure*}[htbp]
    \centering
    \includegraphics[height=\textheight]{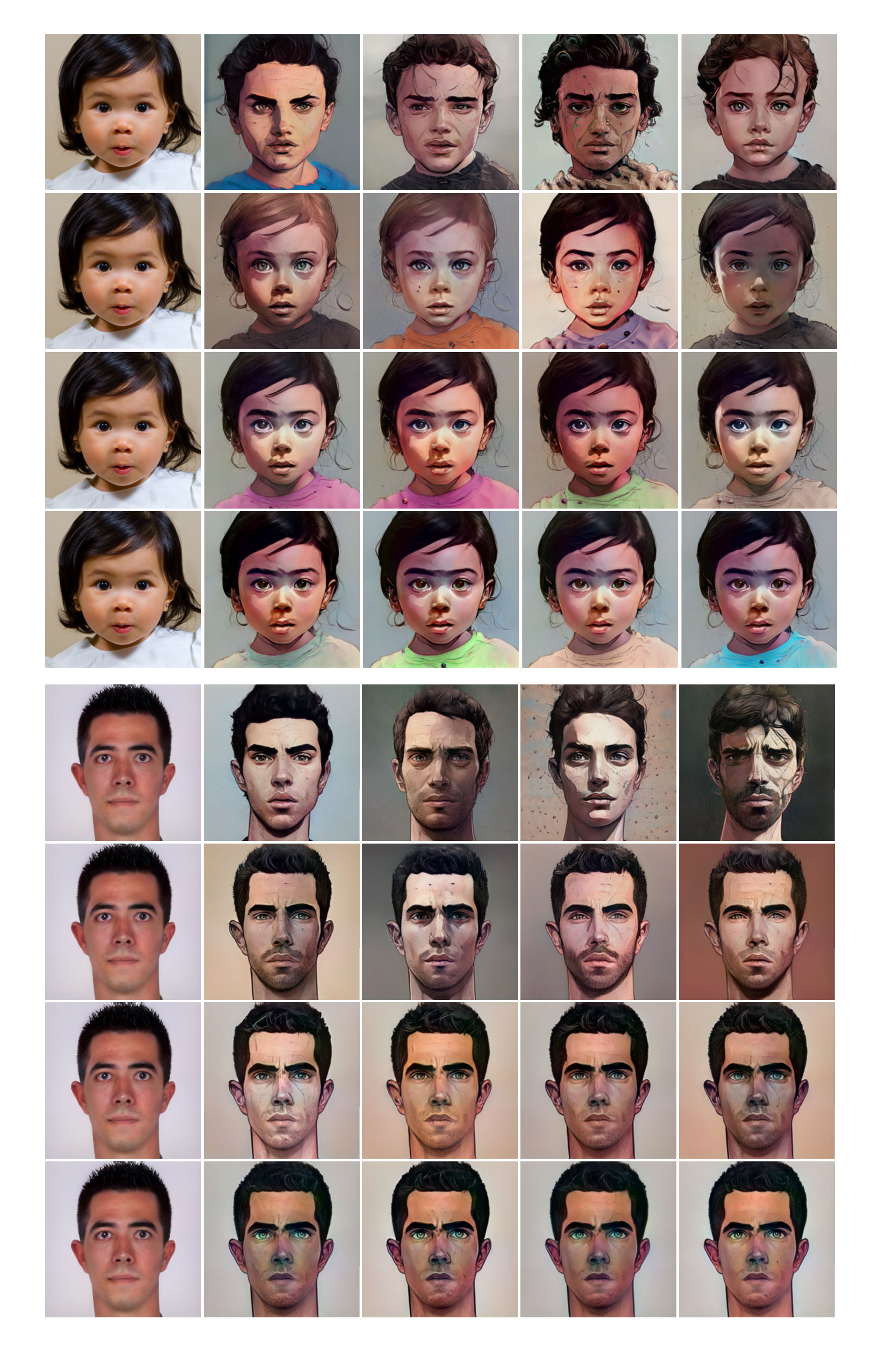}
    \caption{Multimodal portrait stylization results in illustration style.}
    \label{illust-m}
\end{figure*}

\begin{figure*}[htbp]
    \centering
    \includegraphics[height=\textheight]{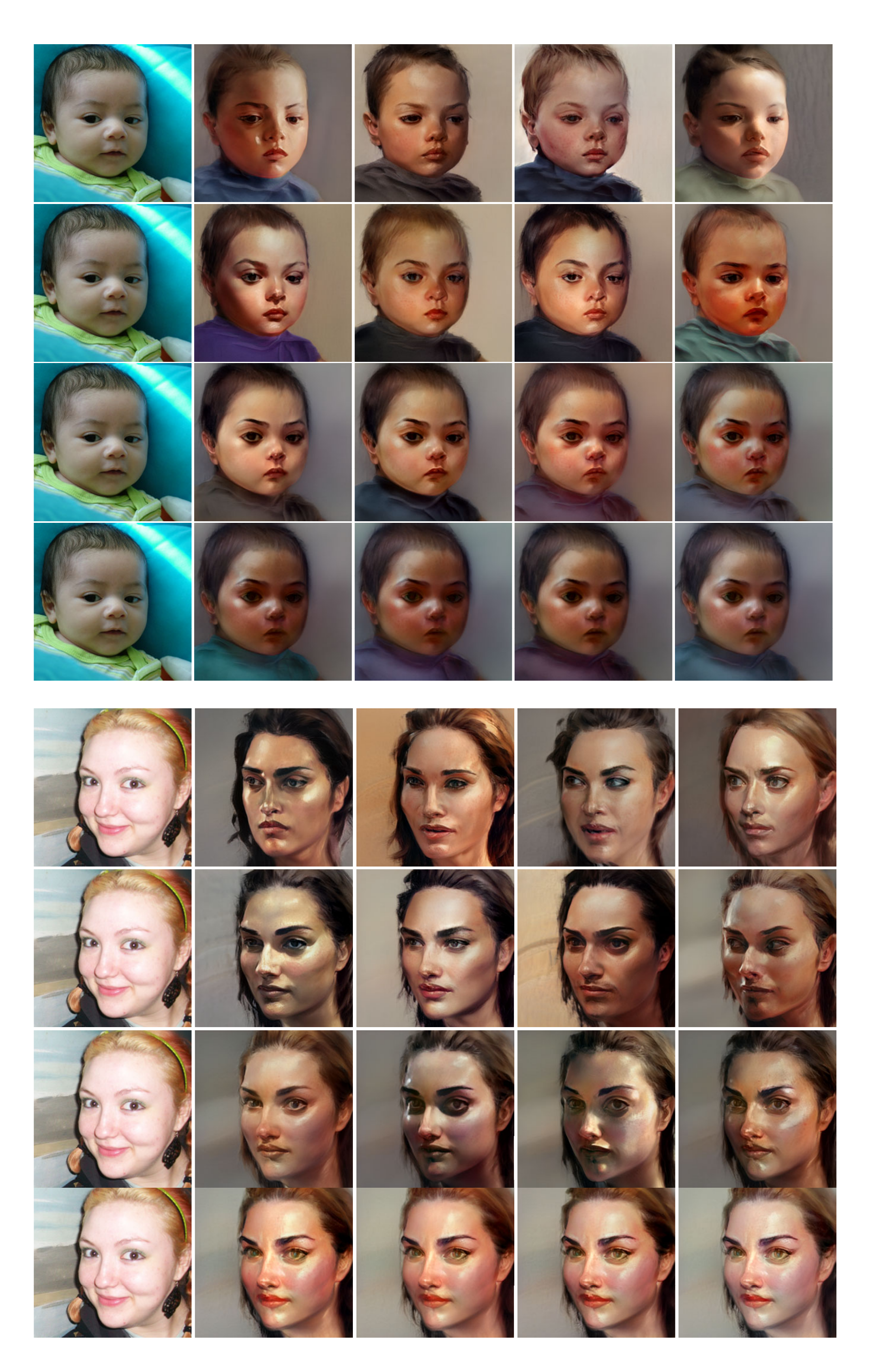}
    \caption{Multimodal portrait stylization results in impasto style.}
    \label{impasto-m}
\end{figure*}

\begin{figure*}[htbp]
    \centering
    \includegraphics[height=\textwidth]{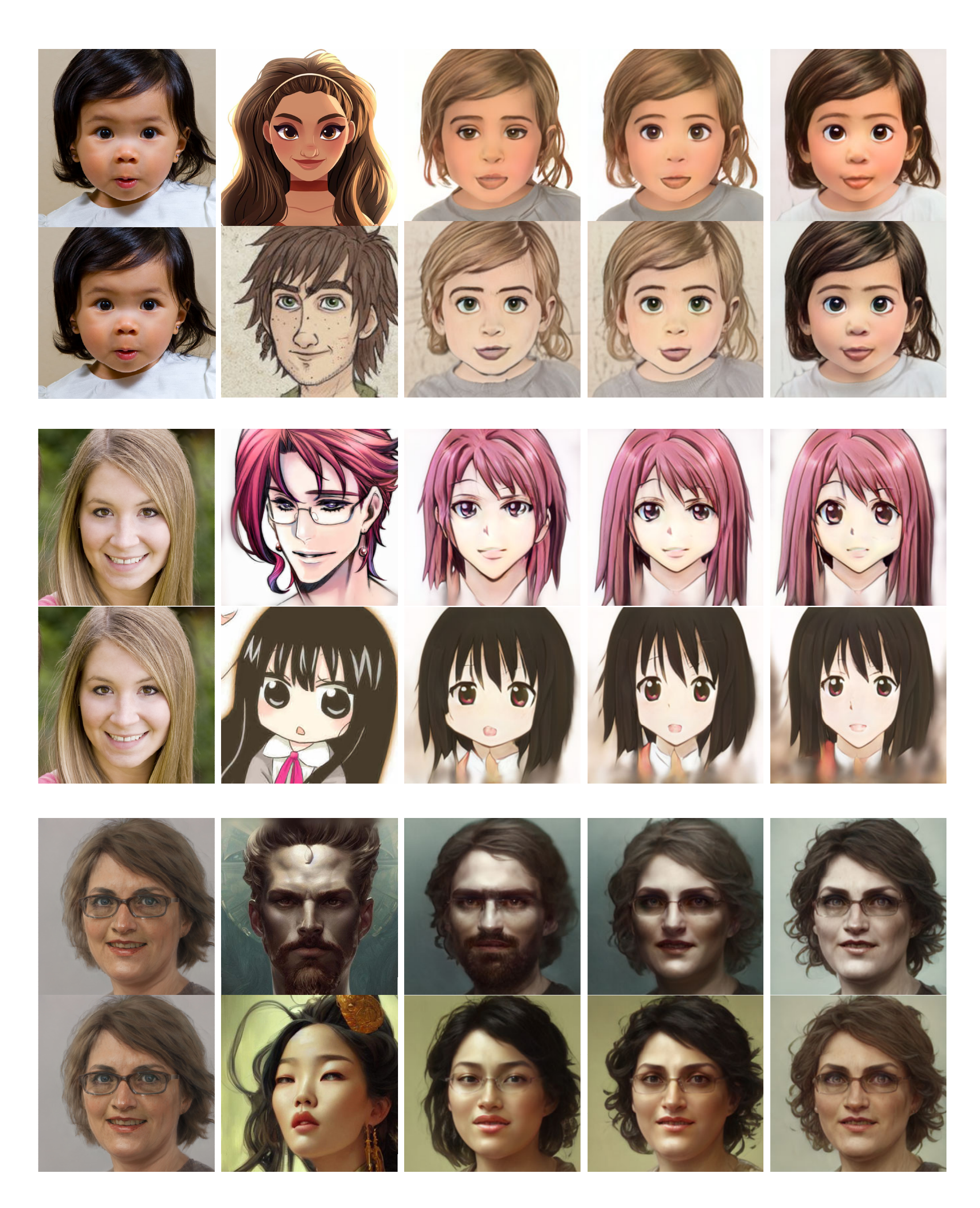}
    \caption{Reference-guided portrait stylization results, with encoding combinations 3, 6 and 9.}
    \label{ref1}
\end{figure*}

\begin{figure*}[htbp]
    \centering
    \includegraphics[width=\textwidth]{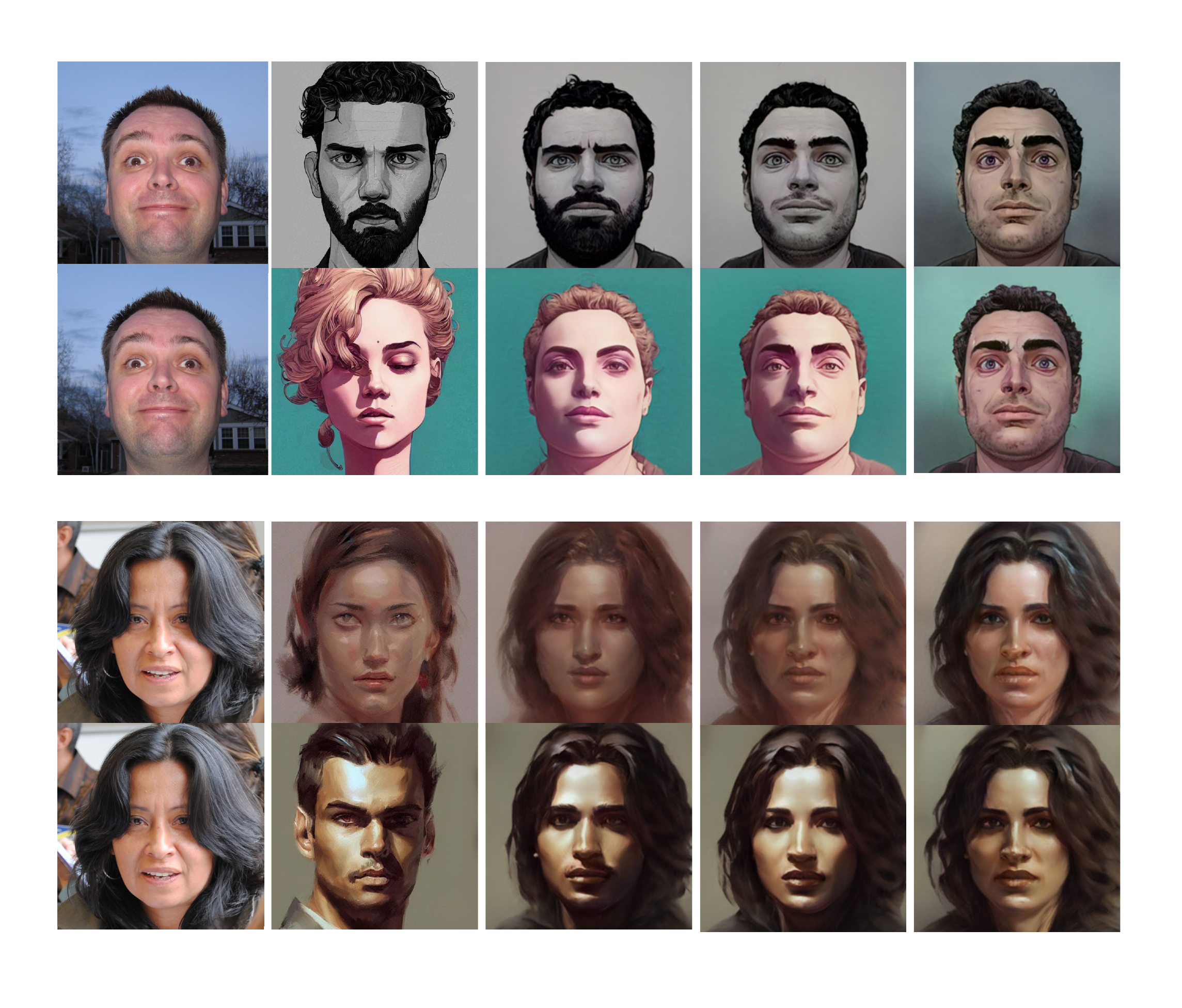}
    \caption{Reference-guided portrait stylization results.}
    \label{ref2}
\end{figure*}

\begin{figure*}[htbp]
    \centering
    \includegraphics[height=\textheight]{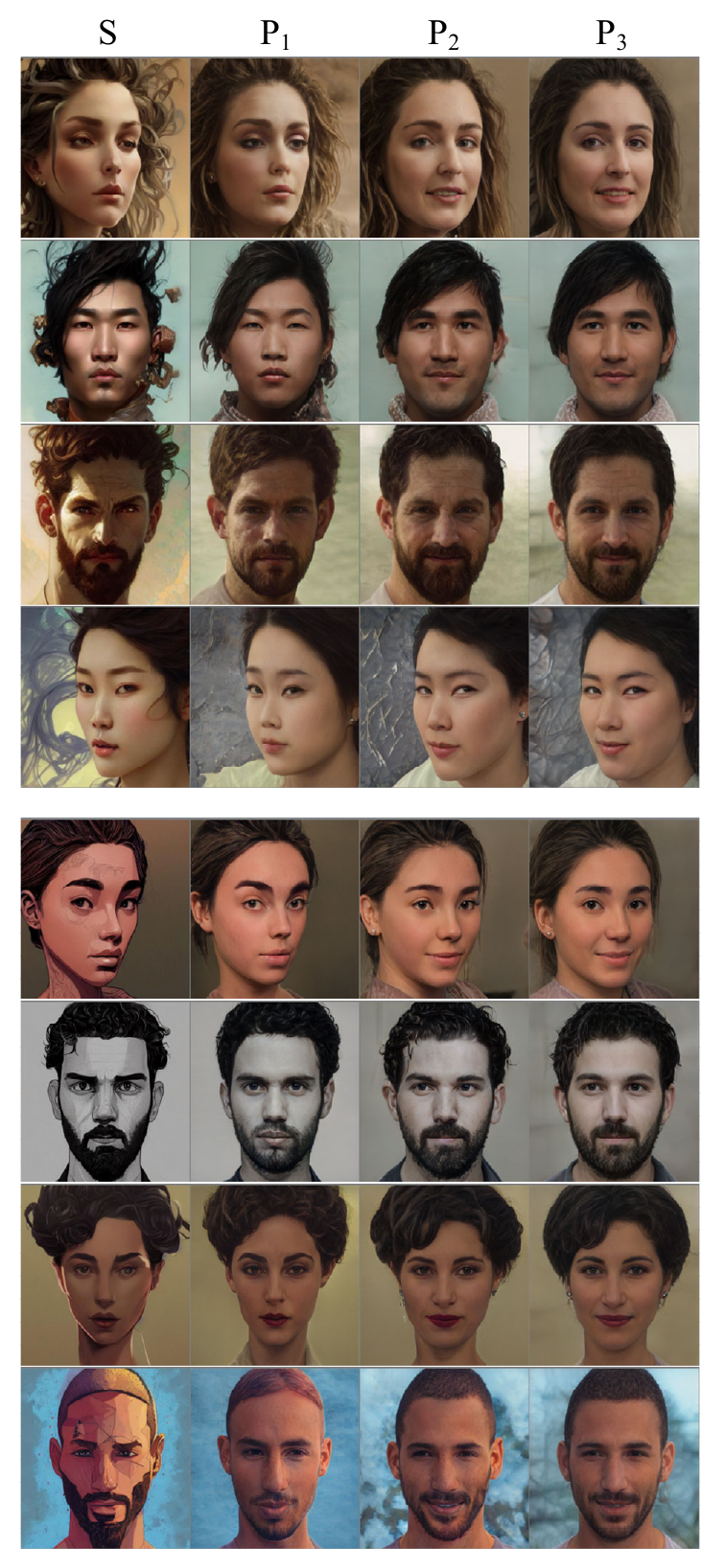}
    \caption{Pseudo-paired dataset: fantasy and illustration style.}
    \label{data1}
\end{figure*}

\begin{figure*}[htbp]
    \centering
    \includegraphics[height=\textheight]{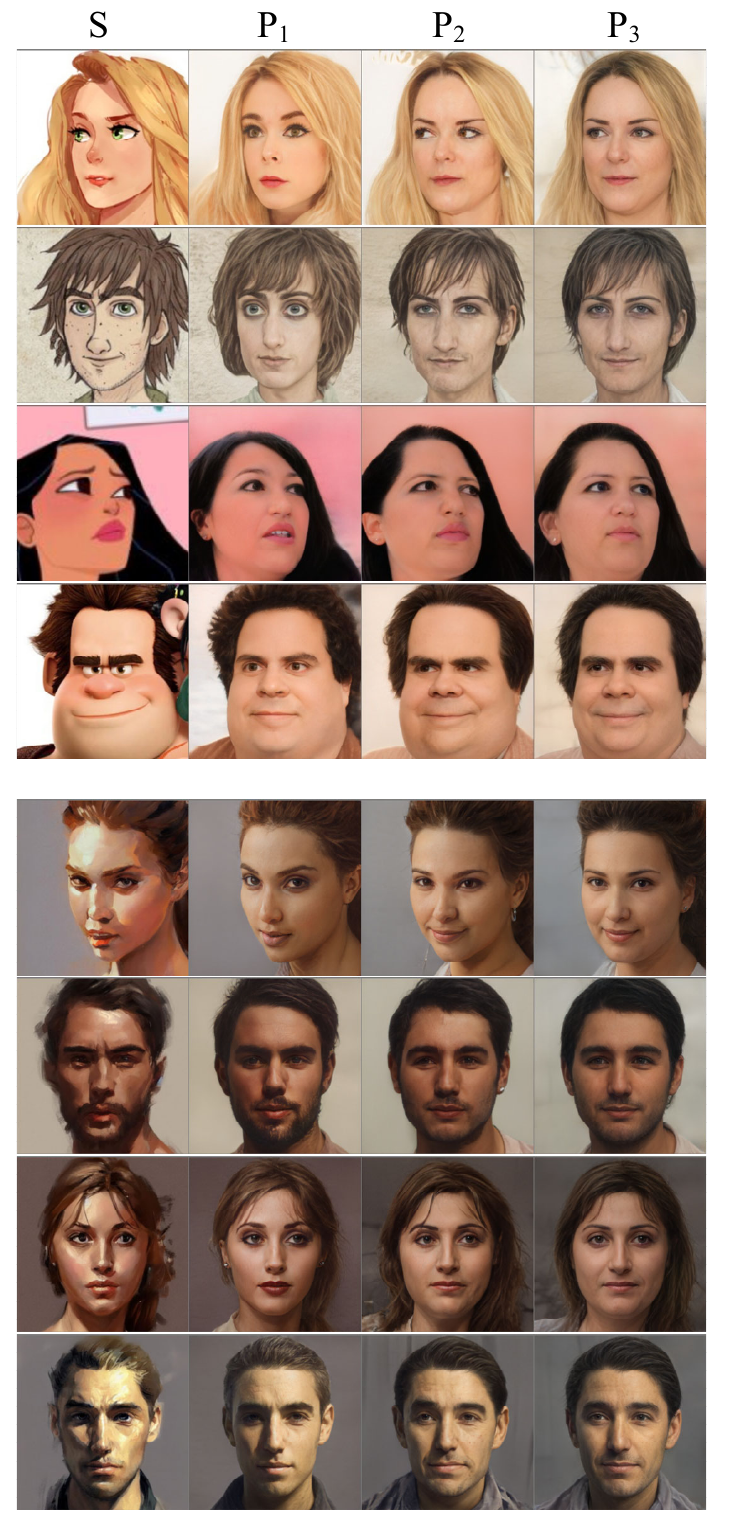}
    \caption{Pseudo-paired dataset: cartoon and impasto style.}
    \label{data2}
\end{figure*}

\begin{figure*}[htbp]
    \centering
    \includegraphics[width=.6\textwidth]{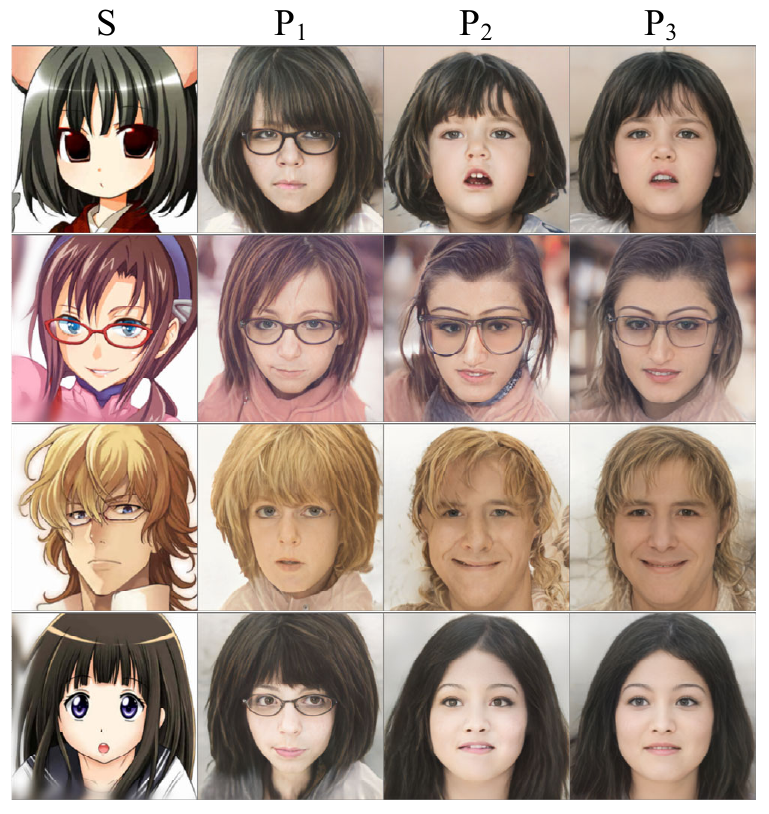}
    \caption{Pseudo-paired dataset: anime style.} 
    \label{data3}
\end{figure*}

More results are provided as follows: portrait stylization results in Figure \ref{normal-1} and \ref{normal-2}; multimodal portrait stylization for each style results in Figures \ref{c-m}, \ref{anime-m}, \ref{fantasy-m}, \ref{illust-m}, and \ref{impasto-m}, respectively; reference-guided portrait stylization in Figures \ref{ref1} and \ref{ref2}. Pseudo-paired data for different styles in Figures \ref{data1}, \ref{data2}, and \ref{data3}.

\subsection{Limitations and Future Work} 

Despite achieving higher quality, more faithful, and flexible portrait stylization, our method has certain limitations.

First, we observe local subtle artifacts in anime styles, likely due to the significant discrepancy between the anime and real domains, as well as insufficient diversity in the style dataset. A more diverse anime dataset could mitigate this issue. Additionally, to reduce computational burden, we applied constraints only to the generator. End-to-end supervision of both the encoder and generator might help reduce these artifacts.

Second, although current diffusion-based methods cannot achieve high-level semantic guidance from reference images for portrait stylization, we still aim to explore their strong potential representation capabilities. The key challenge is appropriately encoding reference images to guide the diffusion model in generating accurate texture and geometric transformations.

From an application perspective, while our method outperforms StyleGAN-based approaches in real-time performance, there remains room for improvement. Achieving high-definition portrait stylization in real time requires a lightweight network. Future research will focus on addressing these challenges.

\bibliographystyle{unsrt}  
\bibliography{references}  

\begin{thebibliography}{10}

\bibitem{cartoon_survey}
Yang Zhao, Diya Ren, Yuan Chen, Wei Jia, Ronggang Wang, and Xiaoping Liu.
\newblock Cartoon image processing: a survey.
\newblock {\em International Journal of Computer Vision}, 130(11):2733--2769, 2022.

\bibitem{cyclegan}
Jun-Yan Zhu, Taesung Park, Phillip Isola, and Alexei~A Efros.
\newblock Unpaired image-to-image translation using cycle-consistent adversarial networks.
\newblock In {\em Proceedings of the IEEE international conference on computer vision}, pages 2223--2232, 2017.

\bibitem{ugatit}
J~Kim.
\newblock U-gat-it: unsupervised generative attentional networks with adaptive layer-instance normalization for image-to-image translation.
\newblock {\em arXiv preprint arXiv:1907.10830}, 2019.

\bibitem{GNR}
Min~Jin Chong and David Forsyth.
\newblock Gans n'roses: Stable, controllable, diverse image to image translation (works for videos too!).
\newblock {\em arXiv preprint arXiv:2106.06561}, 2021.

\bibitem{stylegan}
Tero Karras, Samuli Laine, and Timo Aila.
\newblock A style-based generator architecture for generative adversarial networks.
\newblock In {\em Proceedings of the IEEE/CVF conference on computer vision and pattern recognition}, pages 4401--4410, 2019.

\bibitem{stylegan2}
Tero Karras, Samuli Laine, Miika Aittala, Janne Hellsten, Jaakko Lehtinen, and Timo Aila.
\newblock Analyzing and improving the image quality of stylegan.
\newblock In {\em Proceedings of the IEEE/CVF conference on computer vision and pattern recognition}, pages 8110--8119, 2020.

\bibitem{stylealign}
Zongze Wu, Yotam Nitzan, Eli Shechtman, and Dani Lischinski.
\newblock Stylealign: Analysis and applications of aligned stylegan models.
\newblock {\em arXiv preprint arXiv:2110.11323}, 2021.

\bibitem{toonify}
Justin~NM Pinkney and Doron Adler.
\newblock Resolution dependent gan interpolation for controllable image synthesis between domains.
\newblock {\em arXiv preprint arXiv:2010.05334}, 2020.

\bibitem{UI2I-style}
Jialu Huang, Jing Liao, and Sam Kwong.
\newblock Unsupervised image-to-image translation via pre-trained stylegan2 network.
\newblock {\em IEEE Transactions on Multimedia}, 24:1435--1448, 2021.

\bibitem{agliegan}
Guoxian Song, Linjie Luo, Jing Liu, Wan-Chun Ma, Chunpong Lai, Chuanxia Zheng, and Tat-Jen Cham.
\newblock Agilegan: stylizing portraits by inversion-consistent transfer learning.
\newblock {\em ACM Transactions on Graphics (TOG)}, 40(4):1--13, 2021.

\bibitem{dualstylegan}
Shuai Yang, Liming Jiang, Ziwei Liu, and Chen~Change Loy.
\newblock Pastiche master: Exemplar-based high-resolution portrait style transfer.
\newblock In {\em Proceedings of the IEEE/CVF Conference on Computer Vision and Pattern Recognition}, pages 7693--7702, 2022.

\bibitem{glide}
Alex Nichol, Prafulla Dhariwal, Aditya Ramesh, Pranav Shyam, Pamela Mishkin, Bob McGrew, Ilya Sutskever, and Mark Chen.
\newblock Glide: Towards photorealistic image generation and editing with text-guided diffusion models.
\newblock {\em arXiv preprint arXiv:2112.10741}, 2021.

\bibitem{ldm}
Robin Rombach, Jonas Blattmann, Dominik Lorenz, Patrick Esser, and Bj{\"o}rn Ommer.
\newblock High-resolution image synthesis with latent diffusion models.
\newblock In {\em Proceedings of the IEEE/CVF Conference on Computer Vision and Pattern Recognition (CVPR)}, 2022.

\bibitem{diffusioninstyle}
Martin~Nicolas Everaert, Marco Bocchio, Sami Arpa, Sabine S{\"u}sstrunk, and Radhakrishna Achanta.
\newblock Diffusion in style.
\newblock In {\em Proceedings of the IEEE/CVF International Conference on Computer Vision}, pages 2251--2261, 2023.

\bibitem{stylediffusion}
Zhizhong Wang, Lei Zhao, and Wei Xing.
\newblock Stylediffusion: Controllable disentangled style transfer via diffusion models.
\newblock In {\em Proceedings of the IEEE/CVF International Conference on Computer Vision}, pages 7677--7689, 2023.

\bibitem{inst}
Yuxin Zhang, Nisha Huang, Fan Tang, Haibin Huang, Chongyang Ma, Weiming Dong, and Changsheng Xu.
\newblock Inversion-based style transfer with diffusion models.
\newblock In {\em Proceedings of the IEEE/CVF conference on computer vision and pattern recognition}, pages 10146--10156, 2023.

\bibitem{styleid}
Jiwoo Chung, Sangeek Hyun, and Jae-Pil Heo.
\newblock Style injection in diffusion: A training-free approach for adapting large-scale diffusion models for style transfer.
\newblock In {\em Proceedings of the IEEE/CVF Conference on Computer Vision and Pattern Recognition}, pages 8795--8805, 2024.

\bibitem{GANs}
Ian Goodfellow, Jean Pouget-Abadie, Mehdi Mirza, Bing Xu, David Warde-Farley, Sherjil Ozair, Aaron Courville, and Yoshua Bengio.
\newblock Generative adversarial nets.
\newblock {\em Advances in neural information processing systems}, 27, 2014.

\bibitem{dcgan}
Alec Radford.
\newblock Unsupervised representation learning with deep convolutional generative adversarial networks.
\newblock {\em arXiv preprint arXiv:1511.06434}, 2015.

\bibitem{pix2pix}
Phillip Isola, Jun-Yan Zhu, Tinghui Zhou, and Alexei~A Efros.
\newblock Image-to-image translation with conditional adversarial networks.
\newblock In {\em Proceedings of the IEEE conference on computer vision and pattern recognition}, pages 1125--1134, 2017.

\bibitem{cgan}
Mehdi Mirza and Simon Osindero.
\newblock Conditional generative adversarial nets.
\newblock {\em arXiv preprint arXiv:1411.1784}, 2014.

\bibitem{dualgan}
Zili Yi, Hao Zhang, Ping Tan, and Minglun Gong.
\newblock Dualgan: Unsupervised dual learning for image-to-image translation.
\newblock In {\em Proceedings of the IEEE international conference on computer vision}, pages 2849--2857, 2017.

\bibitem{DiscoGAN}
Taeksoo Kim, Moonsu Cha, Hyunsoo Kim, Jung~Kwon Lee, and Jiwon Kim.
\newblock Learning to discover cross-domain relations with generative adversarial networks.
\newblock In {\em International conference on machine learning}, pages 1857--1865. PMLR, 2017.

\bibitem{nst}
Leon~A Gatys, Alexander~S Ecker, and Matthias Bethge.
\newblock Image style transfer using convolutional neural networks.
\newblock In {\em Proceedings of the IEEE conference on computer vision and pattern recognition}, pages 2414--2423, 2016.

\bibitem{texture}
Dmitry Ulyanov, Vadim Lebedev, Andrea Vedaldi, and Victor Lempitsky.
\newblock Texture networks: Feed-forward synthesis of textures and stylized images.
\newblock {\em arXiv preprint arXiv:1603.03417}, 2016.

\bibitem{perceptual}
Justin Johnson, Alexandre Alahi, and Li~Fei-Fei.
\newblock Perceptual losses for real-time style transfer and super-resolution.
\newblock In {\em Computer Vision--ECCV 2016: 14th European Conference, Amsterdam, The Netherlands, October 11-14, 2016, Proceedings, Part II 14}, pages 694--711. Springer, 2016.

\bibitem{adain}
Xun Huang and Serge Belongie.
\newblock Arbitrary style transfer in real-time with adaptive instance normalization.
\newblock In {\em Proceedings of the IEEE international conference on computer vision}, pages 1501--1510, 2017.

\bibitem{adaain}
Dae~Young Park and Kwang~Hee Lee.
\newblock Arbitrary style transfer with style-attentional networks.
\newblock In {\em proceedings of the IEEE/CVF conference on computer vision and pattern recognition}, pages 5880--5888, 2019.

\bibitem{dynamic}
Yongcheng Jing, Xiao Liu, Yukang Ding, Xinchao Wang, Errui Ding, Mingli Song, and Shilei Wen.
\newblock Dynamic instance normalization for arbitrary style transfer.
\newblock In {\em Proceedings of the AAAI conference on artificial intelligence}, volume~34, pages 4369--4376, 2020.

\bibitem{artflow}
Jie An, Siyu Huang, Yibing Song, Dejing Dou, Wei Liu, and Jiebo Luo.
\newblock Artflow: Unbiased image style transfer via reversible neural flows.
\newblock In {\em Proceedings of the IEEE/CVF Conference on Computer Vision and Pattern Recognition}, pages 862--871, 2021.

\bibitem{styleformer}
Xiaolei Wu, Zhihao Hu, Lu~Sheng, and Dong Xu.
\newblock Styleformer: Real-time arbitrary style transfer via parametric style composition.
\newblock In {\em Proceedings of the IEEE/CVF International Conference on Computer Vision}, pages 14618--14627, 2021.

\bibitem{stytr2}
Yingying Deng, Fan Tang, Weiming Dong, Chongyang Ma, Xingjia Pan, Lei Wang, and Changsheng Xu.
\newblock Stytr2: Image style transfer with transformers.
\newblock In {\em Proceedings of the IEEE/CVF conference on computer vision and pattern recognition}, pages 11326--11336, 2022.

\bibitem{vct}
Bin Cheng, Zuhao Liu, Yunbo Peng, and Yue Lin.
\newblock General image-to-image translation with one-shot image guidance.
\newblock In {\em Proceedings of the IEEE/CVF International Conference on Computer Vision}, pages 22736--22746, 2023.

\bibitem{ntc}
Jing Zhao, Heliang Zheng, Chaoyue Wang, Long Lan, Wanrong Huang, and Wenjing Yang.
\newblock Null-text guidance in diffusion models is secretly a cartoon-style creator.
\newblock In {\em Proceedings of the 31st ACM International Conference on Multimedia}, pages 5143--5152, 2023.

\bibitem{gansurvey}
Weihao Xia, Yulun Zhang, Yujiu Yang, Jing-Hao Xue, Bolei Zhou, and Ming-Hsuan Yang.
\newblock Gan inversion: A survey.
\newblock {\em IEEE transactions on pattern analysis and machine intelligence}, 45(3):3121--3138, 2022.

\bibitem{interpreting}
Yujun Shen, Jinjin Gu, Xiaoou Tang, and Bolei Zhou.
\newblock Interpreting the latent space of gans for semantic face editing.
\newblock In {\em Proceedings of the IEEE/CVF conference on computer vision and pattern recognition}, pages 9243--9252, 2020.

\bibitem{ganspace}
Erik H{\"a}rk{\"o}nen, Aaron Hertzmann, Jaakko Lehtinen, and Sylvain Paris.
\newblock Ganspace: Discovering interpretable gan controls.
\newblock {\em Advances in neural information processing systems}, 33:9841--9850, 2020.

\bibitem{stylespace}
Zongze Wu, Dani Lischinski, and Eli Shechtman.
\newblock Stylespace analysis: Disentangled controls for stylegan image generation.
\newblock In {\em Proceedings of the IEEE/CVF conference on computer vision and pattern recognition}, pages 12863--12872, 2021.

\bibitem{styleclip}
Or~Patashnik, Zongze Wu, Eli Shechtman, Daniel Cohen-Or, and Dani Lischinski.
\newblock Styleclip: Text-driven manipulation of stylegan imagery.
\newblock In {\em Proceedings of the IEEE/CVF international conference on computer vision}, pages 2085--2094, 2021.

\bibitem{sefa}
Yujun Shen and Bolei Zhou.
\newblock Closed-form factorization of latent semantics in gans.
\newblock In {\em Proceedings of the IEEE/CVF conference on computer vision and pattern recognition}, pages 1532--1540, 2021.

\bibitem{image2stylegan}
Rameen Abdal, Yipeng Qin, and Peter Wonka.
\newblock Image2stylegan: How to embed images into the stylegan latent space?
\newblock In {\em Proceedings of the IEEE/CVF international conference on computer vision}, pages 4432--4441, 2019.

\bibitem{image2stylegan++}
Rameen Abdal, Yipeng Qin, and Peter Wonka.
\newblock Image2stylegan++: How to edit the embedded images?
\newblock In {\em Proceedings of the IEEE/CVF conference on computer vision and pattern recognition}, pages 8296--8305, 2020.

\bibitem{psp}
Elad Richardson, Yuval Alaluf, Or~Patashnik, Yotam Nitzan, Yaniv Azar, Stav Shapiro, and Daniel Cohen-Or.
\newblock Encoding in style: a stylegan encoder for image-to-image translation.
\newblock In {\em Proceedings of the IEEE/CVF conference on computer vision and pattern recognition}, pages 2287--2296, 2021.

\bibitem{e4e}
Omer Tov, Yuval Alaluf, Yotam Nitzan, Or~Patashnik, and Daniel Cohen-Or.
\newblock Designing an encoder for stylegan image manipulation.
\newblock {\em ACM Transactions on Graphics (TOG)}, 40(4):1--14, 2021.

\bibitem{restyle}
Yuval Alaluf, Or~Patashnik, and Daniel Cohen-Or.
\newblock Restyle: A residual-based stylegan encoder via iterative refinement.
\newblock In {\em Proceedings of the IEEE/CVF international conference on computer vision}, pages 6711--6720, 2021.

\bibitem{perceptual-loss}
Justin Johnson, Alexandre Alahi, and Li~Fei-Fei.
\newblock Perceptual losses for real-time style transfer and super-resolution.
\newblock In {\em Computer Vision--ECCV 2016: 14th European Conference, Amsterdam, The Netherlands, October 11-14, 2016, Proceedings, Part II 14}, pages 694--711. Springer, 2016.

\bibitem{arcface}
Jiankang Deng, Jia Guo, Niannan Xue, and Stefanos Zafeiriou.
\newblock Arcface: Additive angular margin loss for deep face recognition.
\newblock In {\em Proceedings of the IEEE/CVF conference on computer vision and pattern recognition}, pages 4690--4699, 2019.

\bibitem{lpips}
Richard Zhang, Phillip Isola, Alexei~A Efros, Eli Shechtman, and Oliver Wang.
\newblock The unreasonable effectiveness of deep features as a perceptual metric.
\newblock In {\em Proceedings of the IEEE conference on computer vision and pattern recognition}, pages 586--595, 2018.

\bibitem{danbooru2019Portraits}
Gwern Branwen, Anonymous, and Danbooru Community.
\newblock Danbooru2019 portraits: A large-scale anime head illustration dataset.
\newblock \url{https://gwern.net/crop#danbooru2019-portraits}, March 2019.
\newblock Accessed: DATE.

\bibitem{FID}
Martin Heusel, Hubert Ramsauer, Thomas Unterthiner, Bernhard Nessler, and Sepp Hochreiter.
\newblock Gans trained by a two time-scale update rule converge to a local nash equilibrium.
\newblock {\em Advances in neural information processing systems}, 30, 2017.

\bibitem{adam}
D~Kinga, Jimmy~Ba Adam, et~al.
\newblock A method for stochastic optimization.
\newblock In {\em International conference on learning representations (ICLR)}, volume~5, page~6. San Diego, California;, 2015.

\bibitem{vgg}
Karen Simonyan and Andrew Zisserman.
\newblock Very deep convolutional networks for large-scale image recognition.
\newblock In {\em 3rd International Conference on Learning Representations, ICLR 2015, San Diego, CA, USA, May 7-9, 2015, Conference Track Proceedings}, 2015.

\bibitem{alexnet}
Alex Krizhevsky, Ilya Sutskever, and Geoffrey~E Hinton.
\newblock Imagenet classification with deep convolutional neural networks.
\newblock In {\em Advances in neural information processing systems}, volume~25, pages 1097--1105, 2012.

\bibitem{cmmd}
Sadeep Jayasumana, Srikumar Ramalingam, Andreas Veit, Daniel Glasner, Ayan Chakrabarti, and Sanjiv Kumar.
\newblock Rethinking fid: Towards a better evaluation metric for image generation.
\newblock In {\em Proceedings of the IEEE/CVF Conference on Computer Vision and Pattern Recognition}, pages 9307--9315, 2024.

\end{thebibliography}






\end{document}